%% file: neurips_2026.tex
\documentclass{article}

\usepackage[preprint]{neurips_2026}

\usepackage[utf8]{inputenc}
\usepackage[T1]{fontenc}
\usepackage{hyperref}
\usepackage{url}
\usepackage{booktabs}
\usepackage{amsfonts}
\usepackage{amssymb}
\usepackage{amsmath}
\usepackage{nicefrac}
\usepackage{microtype}
\usepackage{xcolor}
\usepackage{graphicx}
\usepackage{subcaption}
\usepackage{wrapfig}
\usepackage{multirow}
\usepackage{enumitem}

\title{Synthetic Pre-Pre-Training Improves Language Model Robustness
       to Noisy Pre-Training Data}

\author{
  \textbf{Xu Guo}$^{1,2,3}$ \quad
  \textbf{Runyu Peng}$^{1}$ \quad
  \textbf{Jian Tong}$^{1}$ \quad
  \textbf{Yunhua Zhou}$^{1}$ \\
  \textbf{Haijun Lv}$^{1}$ \quad
  \textbf{Zhihui Lu}$^{\dagger 3}$ \quad
  \textbf{Qipeng Guo}$^{\dagger 1,2}$ \\[6pt]
  $^{1}$Shanghai AI Laboratory \quad
  $^{2}$Shanghai Innovation Institute \quad
  $^{3}$Fudan University \\[4pt]
  {\small \texttt{\{pengrunyu, tongjian, zhouyunhua, lvhaijun, guoqipeng\}@pjlab.org.cn}} \\
  {\small \texttt{guox24@m.fudan.edu.cn} \quad \texttt{lzh@fudan.edu.cn}} \\[2pt]
  {\small $^\dagger$Corresponding authors}
}

\begin{document}

\maketitle

%% ============================================================
\begin{abstract}
Large language models (LLMs) rely on web-scale corpora for pre-training. The noise inherent in these datasets tends to obscure meaningful patterns and ultimately degrade model performance.
Data curation mitigates but cannot eliminate such noise, so pre-training corpora remain noisy in practice.
We therefore study whether a lightweight pre-pre-training (PPT) stage based on synthetic data with learnable temporal structure helps resist noisy data during the pre-training (PT) stage.
% Concretely, we draw PPT sequences from many randomly initialized RNNs and keep the PT recipe unchanged.
Across various corruption settings, our method consistently improves robustness to noise during PT, with larger relative gains at higher noise levels.
% The gains generalize across corruption types and naturally noisy web corpora. 
For a 1B-parameter model, a synthetic PPT stage with only 65M tokens achieves the same final loss as the baseline while using up to 49\% fewer natural-text PT tokens across different noise levels.
Mechanistic analyses suggest PPT does not immediately suppress attention to noisy tokens. Rather, PPT-initialized models gradually downweight attention between corrupted tokens during noisy PT. This indicates that synthetic PPT inhibits noise self-modeling and shapes the subsequent optimization trajectory. 
Code is available at \url{https://github.com/guox18/formal-language-prepretraining}.
\end{abstract}

%% ============================================================
\section{Introduction}
\label{sec:intro}
%% ============================================================

Large language models (LLMs) rely on web-scale corpora for pre-training~\citep{penedo2023refinedweb,soldaini2024dolmaopencorpustrillion,penedo2024fineweb,li2025datacomplmsearchgenerationtraining,su2025nemotroncctransformingcommoncrawl}. 
Yet such data inevitably contains noise, from duplicated boilerplate~\citep{elazar2023s} to formatting errors~\citep{zhou2024leveraging}.
Investigations show that the noise inherent in these datasets can reduce the model's knowledge capacity, sometimes by up to 20$\times$~\citep{allen2023physics,allen2024physics}. More recent studies further show that such noise can destabilize training, obscure predictive structure and degrade downstream performance~\citep{ru2025reallyfilterrandomnoise,zhang2026empiricalstudynoisydata}.

To mitigate these issues, large-scale pre-training pipelines rely heavily on data curation~\citep{raffel2020exploring,lee2023deduplicating}. 
However, such curation is inherently incomplete. 
For instance, automated filters could conflate useful rare documents with low-quality text~\citep{longpre2023pretrainersguidetrainingdata}. This creates an inherent trade-off between filtering out noise and preserving tail knowledge.
Furthermore, exhaustive cleaning is computationally prohibitive at web scale~\citep{joulin2016fasttext,albalak2024survey}. This means that pre-training corpora inevitably retain some residual noise. 
A complementary strategy is to enhance the model's intrinsic robustness to noise during training. 
While robust learning from noisy labels is well-established in fields like computer vision~\citep{song2022learning}, improving the noise tolerance of LLMs during pre-training remains largely underexplored~\citep{ru2025reallyfilterrandomnoise}.

Studies show that pre-trained models are more robust to noisy labels than their from-scratch counterparts~\citep{hendrycks2019usingpretrainingimprovemodel}. This suggests that randomly initialized networks are vulnerable to noise. However, this paradigm relies on pre-training to mitigate noise during downstream fine-tuning, and therefore cannot be applied to our scenario. Fortunately, recent work shows that a brief initial training phase on non-natural data (e.g., formal languages) can effectively instill a structural prior in LLMs~\citep{papadimitriou2023injecting}.
However, such non-natural data is typically narrow and homogeneous~\citep{shinnick2025transformers}; this contrasts with natural pre-training data, where both meaningful patterns and noise arise from heterogeneous sources.

Building on this insight, we explore whether a pre-pre-training (PPT) stage can enhance robustness to pre-training (PT) noise. 
Specifically, we introduce a lightweight, synthetic PPT task where the model learns from sequences generated by an ensemble of randomly initialized recurrent neural networks (RNNs). 
Even without training, RNNs inherently produce sequences with temporal structure~\citep{siegelmann1992computational,jaeger2001echo,merrill2020formal}. 
Aggregating outputs from many such independent generators exposes the model to a diverse set of structural patterns. 
This establishes a broad prior for sequential dependencies, guiding the model to prioritize structured signals over unstructured noise during subsequent PT.

\begin{figure*}[t]
\centering
\includegraphics[width=0.95\textwidth]{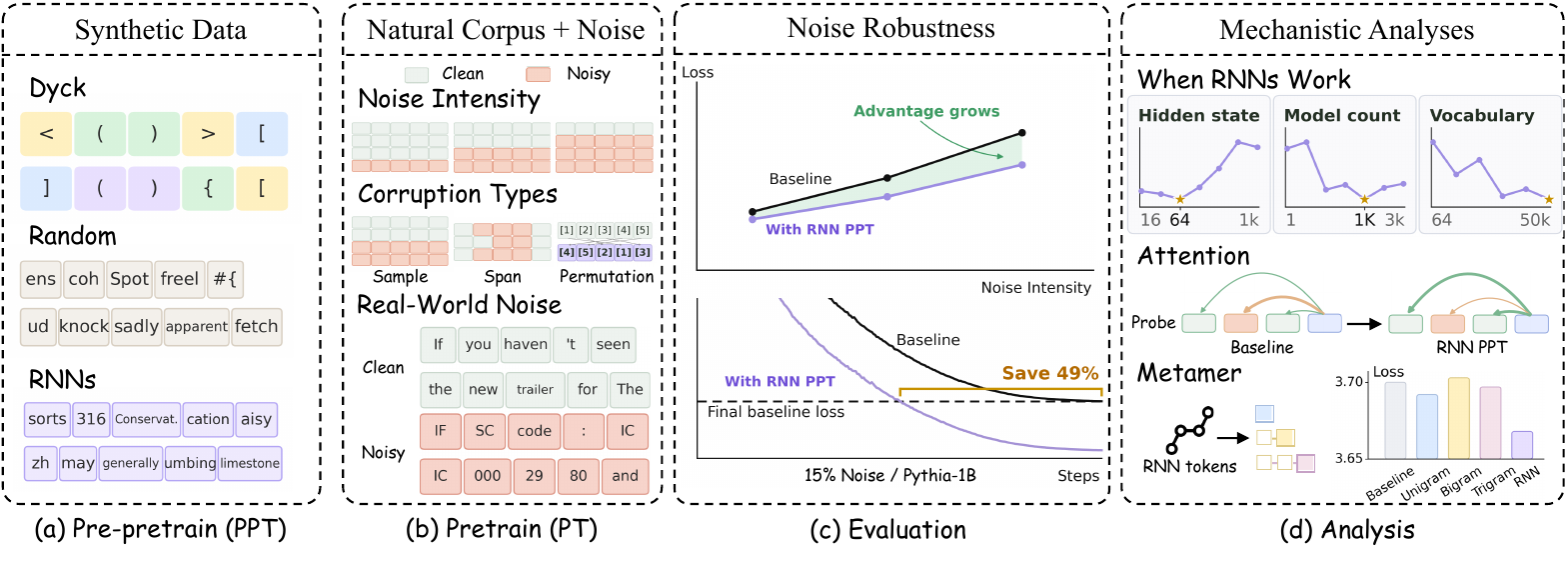}
\caption{Overview of our setup and main findings. (a) We first run a lightweight PPT stage, drawing sequences from one of three synthetic sequences: Dyck, Random, or randomly initialized RNNs. (b) The model is then pre-trained on natural text under controlled noise, varying both intensity and corruption type, and further evaluated on real-world noise. (c) Top: final loss versus noise intensity. Bottom: training loss over steps. (d) We study when RNN-based PPT works and probe its mechanism through attention analyses and metamer controls. These controls match the unigram, bigram, or trigram statistics of RNN-generated tokens.}
\label{fig:intro_teaser}
\end{figure*}

This approach is guided by two design principles, which we ablate in Section~\ref{sec:rnn_ablation}. 
First, the source should be \emph{learnable} for the downstream model, achieved here by using RNNs with moderate hidden state sizes. 
Second, the source should be \emph{less biased}, motivating our use of a broad ensemble and the full vocabulary to generate such RNN sequences. 
Figure~\ref{fig:intro_teaser} summarizes our setup and main findings.

As noise-robust pre-training for LLMs remains underexplored and lacks established baselines, we compare our method against two representative alternatives: an unstructured \emph{Random} i.i.d.\ token source and a formal-language \emph{Dyck} source. 
Across these comparisons, our method consistently outperforms both alternatives, substantially improving model robustness to various types of noise, ranging from controlled corruptions to naturally noisy web data.

The contributions of this paper are threefold:
\begin{itemize}[nosep,leftmargin=*]
\item \textbf{A synthetic PPT recipe for noise-robust pre-training.}
We show that a short PPT stage on random RNN sequences markedly improves the model's noise tolerance during pre-training. The relative gains grow with noise intensity, and transfer across diverse corruption types and naturally noisy web corpora. At the 1B parameter scale with 15\% noise, a 65M-token PPT stage matches the baseline final loss using 49\% fewer PT tokens.
\item \textbf{Design choices that make our RNN-PPT effective.}
Through controlled ablations, we identify three conditions under which RNN-PPT yields gains: generators should remain learnable, the ensemble should be large enough to avoid idiosyncratic bias, and the vocabulary should be broad. Together, these conditions reduce RNN-PPT design to a small set of actionable design choices.
\item \textbf{Mechanistic characterization of the robustness gain.}
We find that the gain emerges during noisy PT, not at PPT initialization. An attention probe shows that RNN-PPT models progressively learn to downweight attention between corrupted tokens---an effect concentrated in late-layer heads. Unigram, bigram, and trigram metamer controls recover almost none of the benefit, indicating that longer-range sequential structure is what drives the effect.
\end{itemize}

%% ============================================================
\section{Related Work}
\label{sec:related}
%% ============================================================
%%
\subsection{Noise in Pre-Training}
Web-scale pre-training corpora are unavoidably noisy~\citep{elazar2023s,zhou2024leveraging}, which reduces knowledge capacity and degrades downstream performance~\citep{allen2023physics,allen2024physics,zhang2026empiricalstudynoisydata}. Curation pipelines~\citep{raffel2020exploring,wenzek2020ccnet,rae2022gopher,penedo2023refinedweb,lee2023deduplicating,gunasekar2023textbooks} mitigate but cannot eliminate residual noise: automated filters risk discarding tail knowledge~\citep{longpre2023pretrainersguidetrainingdata}, and exhaustive cleaning is prohibitive at web scale~\citep{joulin2016fasttext,albalak2024survey}. While Noisy Model Learning has been studied extensively in supervised image classification~\citep{song2022learning}, the intrinsic noise tolerance of LLMs during pre-training remains underexplored~\citep{ru2025reallyfilterrandomnoise}; this paper addresses that gap.

\subsection{Pre-Training on Non-Linguistic Data}
Pre-training on non-linguistic data injects structural inductive biases into language models before they see natural text~\citep{papadimitriou2023injecting}. This direction remains largely underexplored, yet the few existing studies offer illuminating insights.
\citet{papadimitriou2020learning,papadimitriou2023injecting} first show that pre-training on non-linguistic data such as music or code can transfer structural priors to natural language. Subsequent work extends this idea to algorithmic data. \citet{hu2025prepretraining} study Dyck-language variants that accelerate natural language modeling, and explain the effect through the Chomsky hierarchy~\citep{delétang2023neuralnetworkschomskyhierarchy}. ~\citet{shinnick2025transformerspretrainedproceduraldata,jiang2026proceduralpretrainingwarminglanguage} suggest that procedural pre-training induces localized, composable structures across attention and MLP components. \citet{bloem2025universalpretraining} examines how pre-training aids multi-task learning, and explains its effects through Solomonoff induction. 
In this paper, we follow this paradigm but shift the focus to exploring its potential for enhancing noise robustness during pre-training.

%% ============================================================
\section{Method}
\label{sec:method}
%% ============================================================
\subsection{Two-Stage Training Pipeline}
Our training pipeline consists of two stages: a short pre-pre-training (PPT) phase on synthetic data for $S_{\text{ppt}}$ steps, followed by standard pre-training (PT) on natural text for $S_{\text{pt}}$ steps. We evaluate models based on their final validation loss on the PT corpus. To isolate the impact of PPT on noise robustness, we keep the PT data and training recipe fixed across all methods, varying only the synthetic data used during PPT and the resulting parameter initialization.

\subsection{RNN Synthetic PPT Source}
Our proposed source samples sequences from an ensemble of randomly initialized RNN models. Because each generator is a fixed recurrent system, the resulting sequences remain noise-like on the surface while containing temporal structure. 
Training on sequences from multiple such generators exposes the model to diverse structural patterns before it encounters noisy natural text. 
Concretely, for each synthetic sequence we choose a generator $g\sim\mathrm{Unif}(\{1,\ldots,M\})$ and initialize $x_0\sim\mathrm{Unif}(\{1,\ldots,V\})$, $h_0^{(g)}=\mathbf{0}$, where $V$ is the tokenizer vocabulary size. We then run the fixed recurrence $h_t^{(g)}=A^{(g)}e_{x_{t-1}}+W^{(g)}h_{t-1}^{(g)}+b^{(g)}$ with logits $\ell_t^{(g)}=C^{(g)}h_t^{(g)}+d^{(g)}$, where $e_{x_{t-1}}\in\{0,1\}^V$ is the one-hot encoding of the previous token. The next token is then sampled as
\begin{equation}
\label{eq:rnn_sampling}
x_t \sim \operatorname{Categorical}\!\left(\operatorname{softmax}\!\left(\ell_t^{(g)}/\tau\right)\right),
\end{equation}
where $\tau$ is the sampling temperature. Unless ablated, we use $M{=}1000$ generators, hidden size $H{=}64$, the full vocabulary ($V{=}50{,}304$), and $\tau{=}1$. All generator parameters are sampled once at random and then kept fixed; Appendix~\ref{app:rnn_generator} gives the initialization details.
We explore the impact of specific design choices (such as the number of generators) in our ablation studies (Section~\ref{sec:ablation}).

\subsection{Noise Injection}
Following prior studies on noisy pre-training~\citep{ru2025reallyfilterrandomnoise,zhang2026empiricalstudynoisydata}, we apply controlled corruption to a curated corpus, using C4 as a relatively clean baseline. Our primary setting introduces sample-level corruption during PT: with probability $p$, each training sequence is replaced by tokens sampled uniformly from the vocabulary, preserving the original batch size, sequence length, and optimizer settings (full protocol in Appendix~\ref{app:details}).

While sample-level corruption is a controlled proxy for real web noise, we broaden our evaluation with two finer-grained variants---token permutation and span corruption, which perturb token segments rather than entire sequences (Figure~\ref{fig:intro_teaser}(b)). 
To study real web noise, we follow standard practices~\citep{ankner2024perplexed} by using an external reference model to partition FineWeb into clean and noisy subsets (details in Appendix~\ref{sec:fineweb_partition}).

%% ============================================================
\section{Experiments}
\label{sec:exp}
%% ============================================================

\subsection{Experimental Setup}
\label{sec:setup}

\paragraph{Datasets.}
Our primary pre-training corpus is C4~\citep{raffel2020exploring}. To evaluate robustness on real web noise, we also use FineWeb~\citep{penedo2024fineweb} as a natural language dataset.

\paragraph{Baselines.}
We compare four PPT settings: no PPT (pre-training from scratch), Random PPT, Dyck PPT, and our RNN PPT. 
Random PPT serves as a structure-free comparison, sampling tokens independently from the full vocabulary for the same PPT steps. 
Dyck PPT uses $k$-Shuffle Dyck, the strongest formal-language source from prior work~\citep{hu2025prepretraining}, which interleaves $k$ independent, balanced 1-Dyck bracket-matching strings. Following prior work, we set $k {=}64$.

\paragraph{Models.}
In our main experiments, we pre-train a 160M-parameter Pythia model~\citep{biderman2023pythia}, which enables multi-seed sweeps over noise levels and ablations. 
To test whether the effect persists at a larger model size, we also train a 1B-parameter model. Appendix~\ref{app:compute} reports the runtime costs for both scales.

\paragraph{Training.}
Unless stated otherwise, we run $S_{\text{ppt}}=500$ steps of PPT before standard PT. Architecture and hyperparameters remain fixed across comparisons. For token-efficiency comparisons, all experiments use packing, so equal steps correspond to equal training tokens.

\paragraph{Evaluation.}
Our main metric is held-out validation loss on the clean PT corpus, following standard practice in studies of pre-training dynamics~\citep{kaplan2020scalinglawsneurallanguage} and noisy pre-training data~\citep{zhang2026empiricalstudynoisydata}. 
We also report \emph{PT-token savings}: the reduction in PT tokens required to match the baseline's final loss. 
As supplementary downstream evidence, we also evaluate on LAMBADA-OpenAI~\citep{paperno2016lambada}, which requires broad discourse context to predict the final word of a passage; these results are reported in Appendix~\ref{app:downstream_eval}. Results are averaged over three random seeds. 
See Appendix~\ref{app:details} for full experimental details.

\subsection{Main Results}
\label{sec:main_results}
\begin{figure*}[t]
\centering
\includegraphics[width=\textwidth]{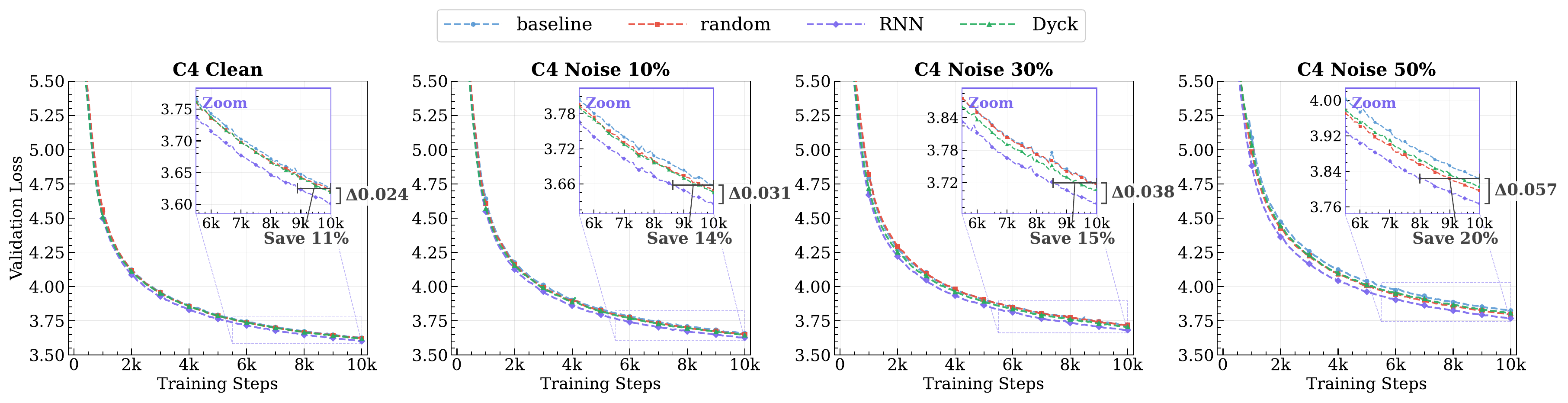}
\caption{Main controlled-noise results on C4, averaged over three seeds. RNN-PPT consistently reaches a lower validation loss, and its advantage grows with the corruption level. Annotations report the final-loss gap and the PT-token savings needed for RNN-PPT to match the baseline. Dyck-PPT shows the same trend but more weakly, and Random PPT closely tracks the baseline.}
\label{fig:main_curves}
\end{figure*}

Figure~\ref{fig:main_curves} shows the controlled-noise results on C4 at 160M scale. RNN-PPT consistently achieves a lower final validation loss than the baseline, with the gap widening at higher noise levels. Across-seed variation is small throughout; for example, in the clean setting, the final-loss standard deviations are $0.001$ for the baseline and $0.002$ for RNN-PPT (Appendix Table~\ref{tab:c4ppt_160m}). This translates to substantial PT-token savings. Dyck-PPT offers modest improvements, while Random PPT performs similarly to the baseline, indicating that structure, not just warm-up, is necessary for robustness.

\subsection{Generalization Across Noise Types}
\label{sec:noise_types}
\begin{figure*}[t]
\centering
\includegraphics[width=0.96\textwidth]{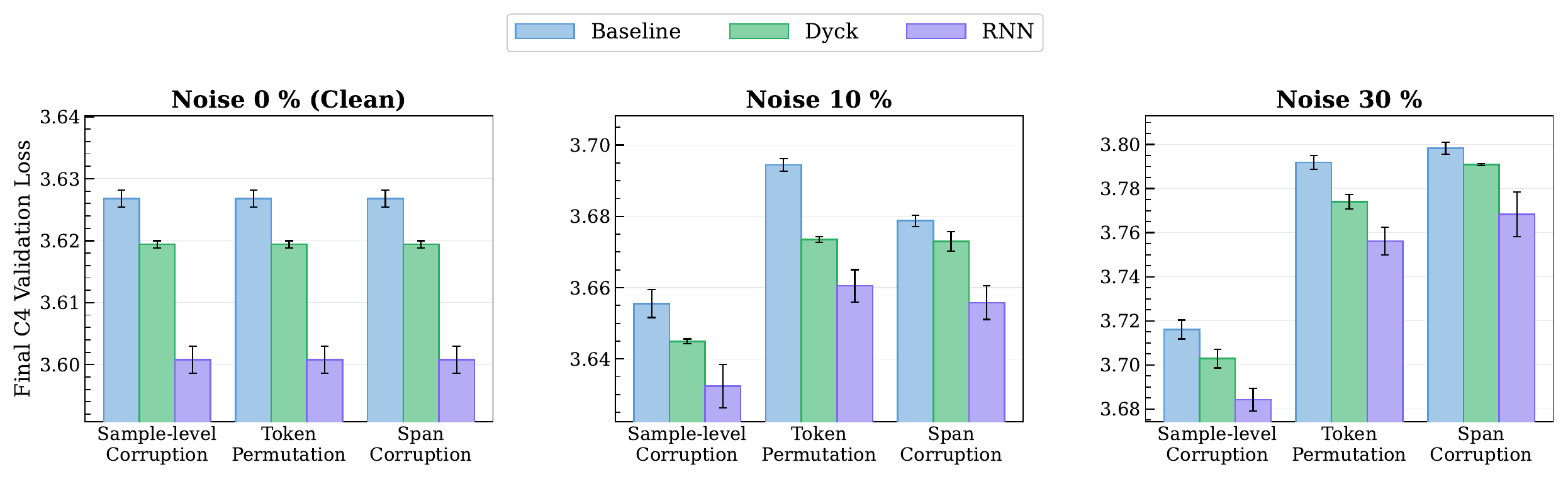}
\caption{Generalization across corruption types. RNN-PPT improves final loss under all three protocols (sample-level corruption, token permutation, and span corruption), showing that the benefit is not tied to any single corruption type. Bars show mean final validation loss over three seeds; error bars denote standard deviation.}
\label{fig:noise_types}
\end{figure*}

We extend our evaluation beyond sample-level corruption to token permutation and span corruption (Figure~\ref{fig:noise_types}). RNN-PPT consistently improves final validation loss across all three noise types, with the most significant gains under token permutation. Dyck-PPT provides smaller benefits. This confirms that RNN-PPT's robustness is agnostic to the specific corruption protocol.

\begin{figure*}[t]
\centering
\begin{minipage}[t]{0.58\textwidth}
\centering
\includegraphics[width=\linewidth]{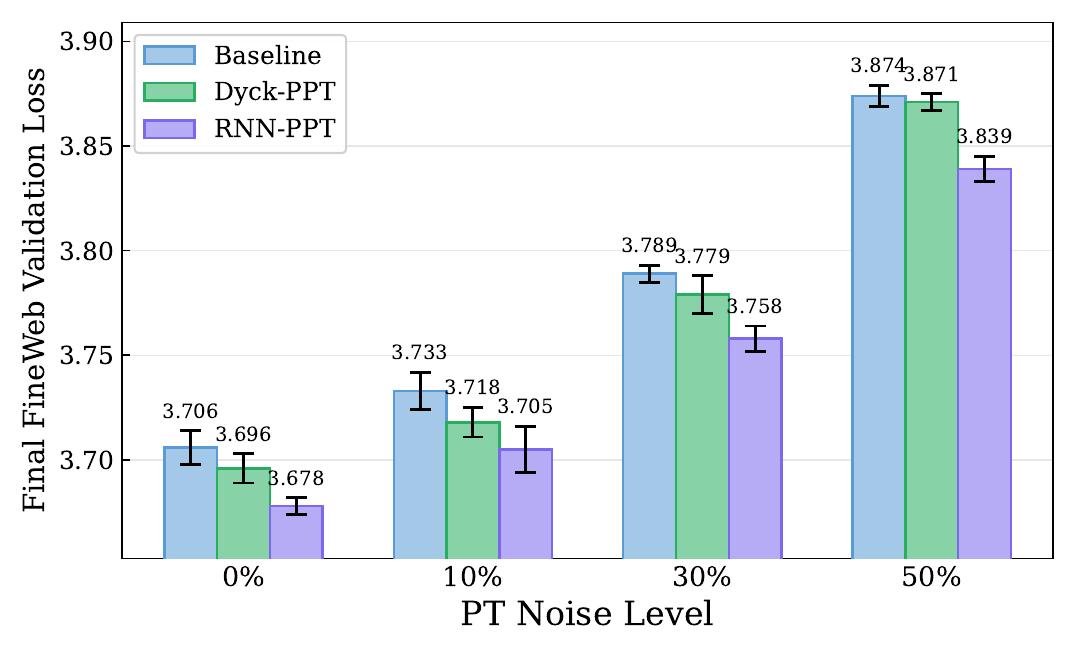}
\captionof{figure}{Controlled-noise FineWeb results. Bars show mean final FineWeb validation loss over three seeds and error bars denote standard deviation. RNN-PPT is best at every tested PT noise level.}
\label{fig:fineweb_controlled}
\end{minipage}
\hfill
\begin{minipage}[t]{0.41\textwidth}
\centering
\includegraphics[width=\linewidth]{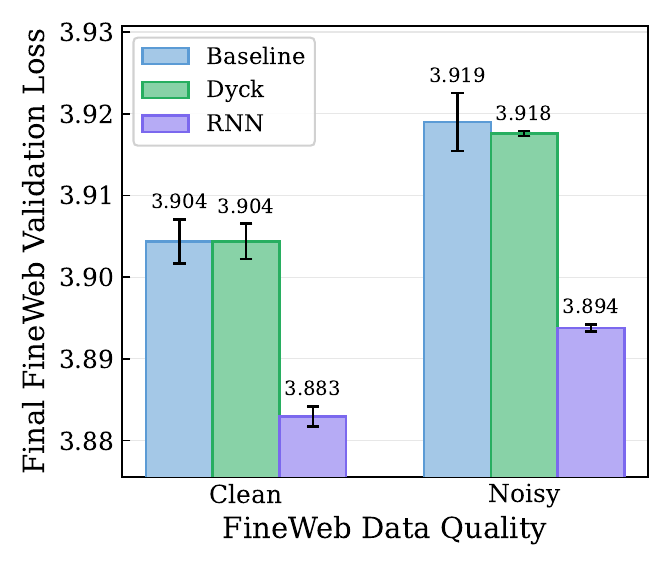}
\captionof{figure}{Validation on naturally noisy data. RNN-PPT lowers validation loss on both quality splits, with a larger gain on the noisier one.}
\label{fig:fineweb_real_noise}
\end{minipage}
\end{figure*}

\subsection{Robustness on FineWeb}
\label{sec:fineweb}
We next test whether the robustness gain transfers beyond C4, first under a controlled FineWeb corruption protocol and then on naturally noisy FineWeb subsets.

\paragraph{Controlled Noise.} We first apply the same controlled-noise protocol to FineWeb (Figure~\ref{fig:fineweb_controlled}). RNN-PPT achieves the lowest final validation loss at every tested noise level, outperforming both the baseline and Dyck-PPT. Thus, the effect is not specific to C4.

\paragraph{Naturally Noisy Data.} We then partition FineWeb into clean and noisy subsets using an external reference model~\citep{ankner2024perplexed} (Appendix~\ref{sec:fineweb_partition}). The noisy subset contains naturally occurring, less informative web fragments; representative cases appear in Appendix~\ref{app:fineweb_cases}. RNN-PPT lowers validation loss on both splits (Figure~\ref{fig:fineweb_real_noise}), with the larger gain on the noisier split. Dyck-PPT has little effect in this setting, suggesting that RNN-PPT transfers more effectively to natural web noise.

\subsection{Robustness Persists at Larger Scale}
\label{sec:scale}
Figure~\ref{fig:scale_1b} tests whether the effect persists at 1B-parameter scale. Since the smaller 160M experiments already cover a broad range of corruption levels, the 1B experiment focuses on a finer-grained and more practical noise range. RNN-PPT consistently outperforms the baseline at every tested noise level, despite the larger model capacity and longer PT duration. Appendix~\ref{app:longpt} shows the same trend when extending the 160M PT budget from 10K to 20K steps.

We also include a budget-matched C4-PPT control to separate the effect of the RNN source from the effect of seeing additional clean, PT-like data. C4-PPT yields only marginal gains at 1B. Appendix~\ref{app:c4ppt_scales} reports the corresponding 160M comparison, where RNN-PPT remains the strongest method. This suggests that the robustness gain is not explained by extra C4 training.

The magnitude of the 1B gain is practically meaningful and consistent across noise.
At the 1B scale, RNN-PPT reduces final validation loss by $0.10$--$0.15$ across the tested noise range. The gain also translates into sample efficiency: after the 65M-token RNN-PPT stage, the model reaches the baseline's final loss using up to $49\%$ fewer natural-text PT tokens.
Beyond held-out validation loss, RNN-PPT also tends to improve LAMBADA~\citep{paperno2016lambada} perplexity; full downstream results are in Appendix~\ref{app:downstream_eval}.

%% ============================================================
\section{Ablation Study}
\label{sec:ablation}
%% ============================================================
In this section, we analyze which parts of the RNN-PPT design contribute to its robustness: the PPT budget (\S\ref{sec:ppt_budget}) and RNN generator choices (\S\ref{sec:rnn_ablation}).

\subsection{PPT Budget}
\label{sec:ppt_budget}
\begin{figure}[t]
\centering
\includegraphics[width=0.9\linewidth]{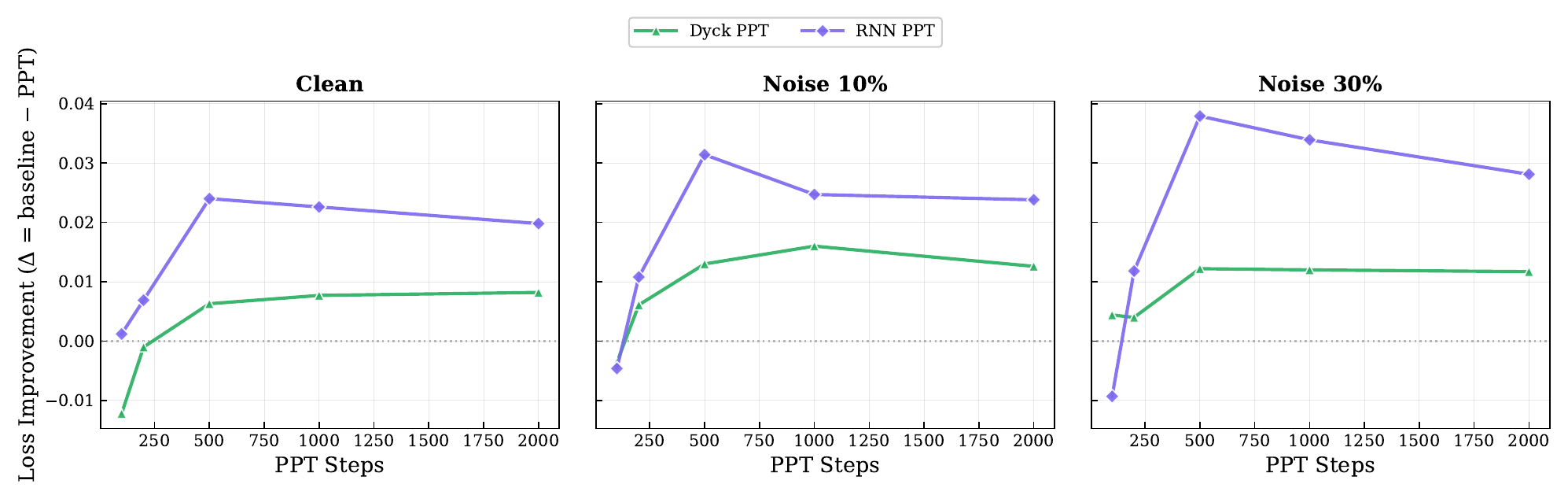}
\caption{Sensitivity to PPT budget. Improvements over the baseline appear after a few hundred synthetic steps and then largely plateau. For RNN-PPT, 500 steps is close to the best PPT budget across noise levels.}
\label{fig:ppt_budget}
\end{figure}

We vary the synthetic warm-up length in Figure~\ref{fig:ppt_budget}. The benefit emerges within a few hundred steps and then plateaus. At $30\%$ noise, RNN-PPT is slightly worse than the baseline at $100$ steps, matches it around $200$ steps, and improves further to peak near $500$ steps; longer PPT brings no further gain. Dyck-PPT exhibits a similar but less pronounced trend. We therefore adopt $500$ PPT steps as the default PPT budget: a lightweight intervention that already captures most of the robustness gain.

\subsection{RNN Generator Design}
\label{sec:rnn_ablation}
\begin{figure*}[t]
\centering
\includegraphics[width=0.96\textwidth]{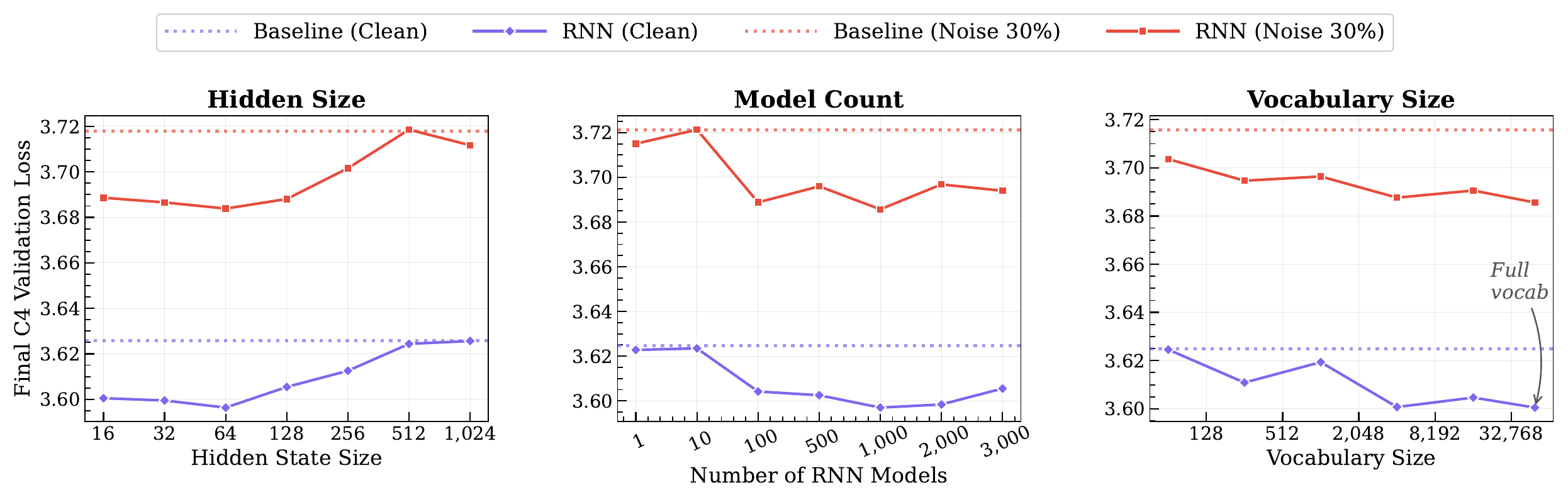}
\caption{RNN design ablations at $0\%$ and $30\%$ PT noise. \textbf{Left:} Transfer is strongest within a moderate generator-complexity range. \textbf{Middle} and \textbf{Right:} Larger ensembles and broader vocabularies yield the best overall robustness, supporting our default of a large ensemble and full vocabulary.}
\label{fig:rnn_ablation}
\end{figure*}

Section~\ref{sec:intro} motivated RNN-PPT with two design principles: the synthetic source should be learnable by the downstream model, and it should be less biased. We test learnability by varying RNN hidden size, and test low-bias design by varying ensemble size and vocabulary size. Figure~\ref{fig:rnn_ablation} shows the $0\%$ and $30\%$ PT settings for readability; Appendix~\ref{app:rnn_ablation_noise} reports the complementary $10\%$ and $50\%$ results.

\paragraph{Learnability.}
The hidden-size sweep (Figure~\ref{fig:rnn_ablation}, Left) shows a broad sweet spot. Hidden size $64$ attains the lowest final loss under $30\%$ noise and remains near the best at $10\%$ noise, while the $16$--$64$ range is strongest at $50\%$. With much larger generators ($512$/$1024$ hidden sizes), the gain over the baseline nearly disappears. Thus, RNN-PPT helps when the generated sequences are relatively simple for the downstream model to learn, but not when the generator becomes too complex.

\paragraph{Low-bias source design.}
The generator-count and vocabulary sweeps support the low-bias principle from two angles. For generator count (Figure~\ref{fig:rnn_ablation}, Middle), one or ten generators yield little benefit; gains emerge around $100$ generators and remain effective through $3000$. Appendix~\ref{app:rnn_ablation_noise} shows the same trend at $10\%$ and $50\%$ noise, with $1000$ generators performing best. This suggests that too few generators might transfer idiosyncratic recurrent patterns, while a larger ensemble provides a broader sequential prior.

The vocabulary sweep (Figure~\ref{fig:rnn_ablation}, Right) shows a similar pattern for token support. Restricted vocabularies recover only part of the benefit, whereas larger vocabularies generally work better. Since the full vocabulary performs best in the clean setting and in most noisy settings, we use it as the default. Together, these results suggest that RNN-PPT is most effective when its structure is broad rather than tied to a small generator set or a narrow token subset.

%% ============================================================
\section{Analysis}
\label{sec:analysis}
%% ============================================================
The preceding experiments demonstrate the effectiveness and robustness of RNN-PPT under varied conditions. We now conduct two diagnostic analyses to better understand what the RNN source contributes: attention probes examine whether models pre-trained with noisy data rely on corrupted tokens to predict other corrupted tokens, while metamer controls assess whether low-order token statistics are sufficient to explain the observed transfer gain.

\subsection{Noise Self-Modeling in Attention}
\label{sec:attention}

Models trained from scratch are more vulnerable to noisy data. We hypothesize that this vulnerability arises because they learn to model the noise itself as a predictable pattern, relying on earlier corrupted tokens to process subsequent ones.
We refer to this behavior as \emph{noise self-modeling}.

Specifically, we investigate whether models use noisy tokens to predict other noisy 
tokens under noisy PT. 
We ask (i)~\emph{when} the gap between RNN-PPT and the baseline emerges during PT, and (ii)~\emph{where} in the model this difference is localized. The probe setup is given in Appendix~\ref{app:attention_probe}.

\paragraph{Metric.} We define an attention-based metric to quantify noise self-modeling, which captures the extent to which representations at noisy positions are constructed from preceding corrupted positions in the causal context.

Concretely, let $a_{qk}$ denote the attention weight from query position $q$ to key position $k$. For a given $q$, let $K_{\text{valid}}(q)$ be all valid key positions in its causal prefix, and let $K_{\text{noise}}(q)\subseteq K_{\text{valid}}(q)$ be the noisy ones. Since attention weights are normalized over the valid prefix, $\sum_{k\in K_{\text{valid}}(q)} a_{qk}=1$, the inner sum in Equation~\ref{eq:r_noise} is the fraction of query $q$'s attention mass assigned to noisy keys. We average this quantity over the noisy query set $Q_{\text{noise}}$:
\begin{equation}
\label{eq:r_noise}
r_{\text{noise}}
\;=\;
\frac{1}{\lvert Q_{\text{noise}}\rvert}
\sum_{q\in Q_{\text{noise}}}\;\sum_{k\in K_{\text{noise}}(q)} a_{qk}\,.
\end{equation}
We define $\Delta r_{\text{noise}} = r_{\text{noise}}^{\text{RNN-PPT}} - r_{\text{noise}}^{\text{No-PPT}}$ per (layer, head) and average across three model seeds.

\begin{figure*}[t]
\centering
\begin{minipage}[t]{0.5\textwidth}
\centering
\includegraphics[width=\linewidth]{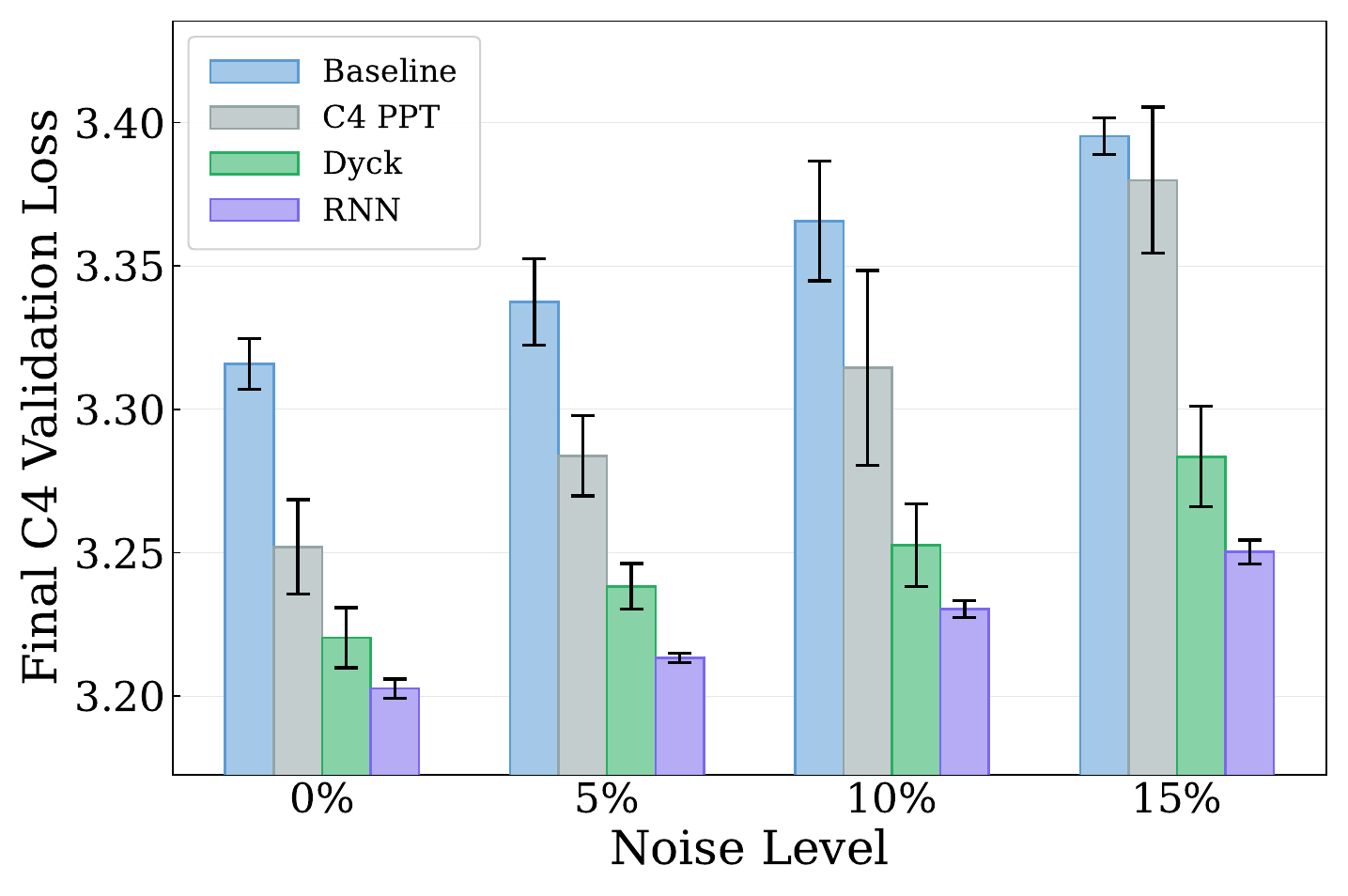}
\captionof{figure}{Method comparison at 1B scale. With a 25K-step PT budget, RNN-PPT remains effective at larger model size across all tested noise levels.}
\label{fig:scale_1b}
\end{minipage}
\hfill
\begin{minipage}[t]{0.49\textwidth}
\centering
\includegraphics[width=\linewidth]{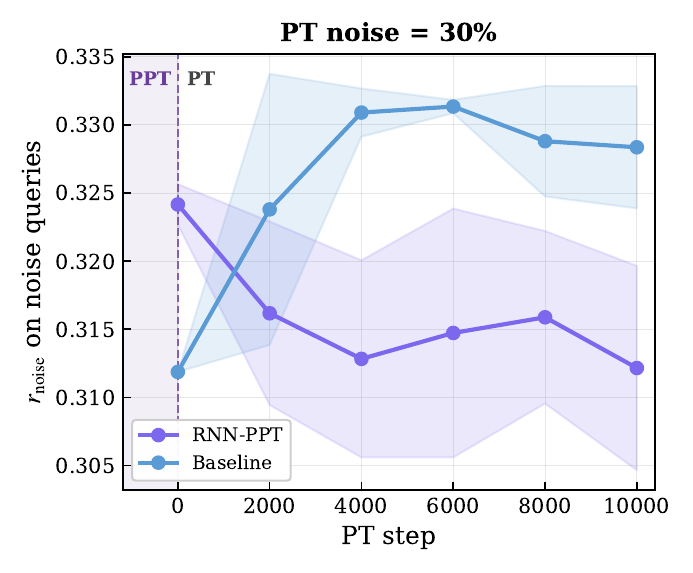}
\captionof{figure}{Layer-mean noise self-modeling, measured as $r_{\text{noise}}$ on noisy queries over PT steps at $30\%$ noise. Shaded band: $\pm$sd across three seeds.}
\label{fig:attention_dynamics}
\end{minipage}
\end{figure*}

\begin{figure*}[t]
\centering
\includegraphics[width=\linewidth]{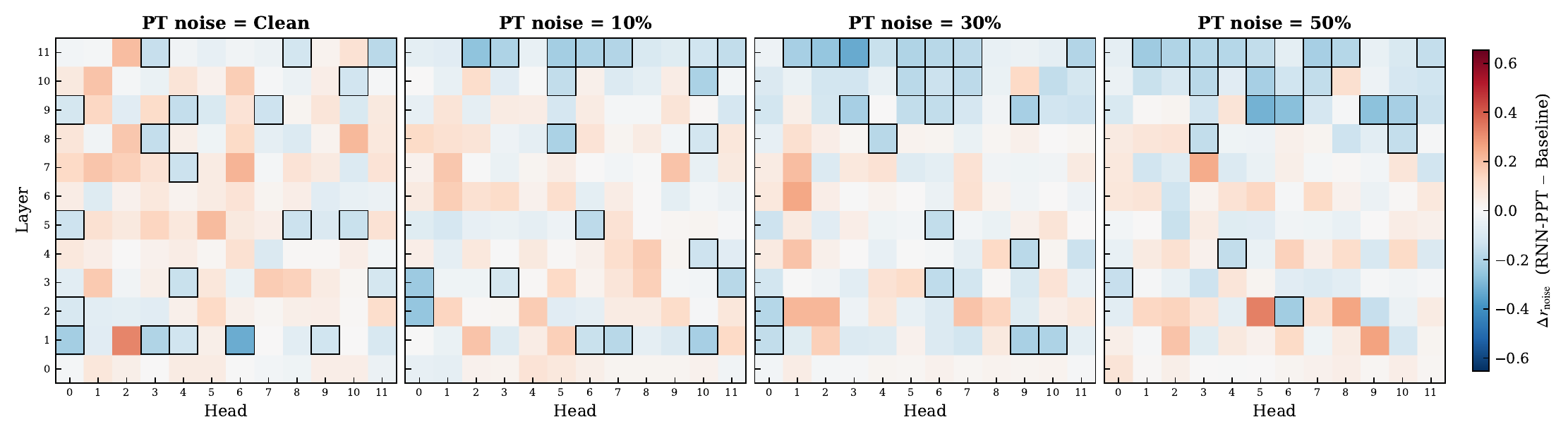}
\captionof{figure}{Seed-mean per-(layer, head) $\Delta r_{\text{noise}}$ on noisy query tokens under the fixed probe setting described in Appendix~\ref{app:attention_probe}. Panels vary the PT noise level from clean to $50\%$. Blue indicates weaker noise self-modeling for RNN-PPT than for models without PPT at that head. Black borders mark the top-$20$ most-negative heads per panel.}
\label{fig:attention_head_heatmap}
\end{figure*}

Figure~\ref{fig:attention_dynamics} tracks the layer-mean $r_{\text{noise}}$ on noisy queries across PT checkpoints. No-PPT models begin from the randomly initialized checkpoint. As noisy PT proceeds, the no-PPT models increasingly attend from noisy query tokens to noisy key tokens. This suggests that the model learns to treat corrupted tokens as predictive of other corrupted tokens. RNN-PPT changes this trajectory: within the first $2$K PT steps, its curve drops below the no-PPT curve, and the gap remains throughout PT. This indicates that RNN-PPT suppresses noise self-modeling during PT. The reduction in noise self-modeling emerges early and persists, rather than appearing as a late-training correction.

Figure~\ref{fig:attention_head_heatmap} visualizes $\Delta r_{\text{noise}}$ per (layer, head) for checkpoints trained with PT noise, using a fixed probe setting. The effect is not spread evenly across heads: the largest reductions in noise self-modeling concentrate in layers~8--11, and intensify as PT noise rises. 
The effect is localized rather than uniform, indicating that later layers drive the reduction in noise self-modeling. This suggests PPT mitigates noise during high-level integration rather than at shallow processing stages.

\subsection{Unigram, Bigram, and Trigram Metamer Controls}
\label{sec:metamer}

\begin{wraptable}{r}{0.5\textwidth}
\vspace{-0.8\baselineskip}
\centering
\footnotesize
\setlength{\tabcolsep}{4pt}
\begin{tabular}{lcccc}
\toprule
PPT source & $0\%$ & $10\%$ & $30\%$ & $50\%$ \\
\midrule
Unigram metamer & 0.006 & 0.006 & 0.008 & 0.014 \\
Bigram metamer & $-$0.002 & 0.002 & $-$0.011 & $-$0.001 \\
Trigram metamer & 0.002 & 0.003 & 0.004 & 0.005 \\
RNN subset    & \textbf{0.026} & \textbf{0.029} & \textbf{0.036} & \textbf{0.052} \\
\bottomrule
\end{tabular}
\caption{Unigram, bigram, and trigram metamer diagnostic. Entries report mean improvement over the no-PPT baseline in final validation loss across three seeds.}
\label{tab:metamer}
\vspace{-1.0\baselineskip}
\end{wraptable}

A natural concern is whether RNN-PPT's benefit comes mainly from low-order token statistics rather than from richer sequential structure. 
To test this possibility, we construct unigram, bigram, and trigram metamers from the RNN source subset for 500 PPT steps. 
For each order, we fit the token-level model to the RNN source tokens and then sample new sequences from it, preserving local statistics while removing the original long-range organization. We then re-run the full pipeline with each metamer as the PPT source, holding the PPT budget and PT recipe fixed. 
Table~\ref{tab:metamer} reports the mean improvement over the no-PPT baseline across three seeds; construction details are given in Appendix~\ref{app:metamer}.
These metamers produce only marginal gains. This suggests that RNN-PPT's benefit depends on longer-range dependencies, not simply on matching low-order token statistics.

%% ============================================================
\section{Discussion}
\label{sec:discussion}
%% ============================================================

How does RNN-PPT impart such robustness against noisy data? A natural explanation is that learning synthetic sequential structure during PPT makes the model downweight noise. 
Our attention analyses in Section~\ref{sec:analysis} do not support this.
At the start of PT, RNN-PPT models attend more strongly among corrupted tokens than models without PPT, suggesting that PPT has not taught the model to directly downweight noise. 
The benefit emerges during noisy PT: RNN-PPT models quickly reduce noise-to-noise attention, while models without PPT drift in the opposite direction. 
Thus, RNN-PPT improves robustness not by directly downweighting noisy tokens during PPT, but by guiding the subsequent optimization trajectory toward lower noise self-modeling.

More broadly, PPT acts as a complement rather than a replacement for data curation. Curation intervenes on the corpus, but remains incomplete in practice and can leave residual noise. PPT intervenes on the model: it leaves the corpus and PT recipe untouched, yet makes training more tolerant of noisy mixtures. 

In this work, motivated by the hypothesis in Section~\ref{sec:intro}, we chose randomly initialized RNNs as the synthetic source and validated this choice empirically. We note three limitations that point to natural follow-ups. First, testing each candidate source requires running the full PPT-to-PT pipeline, making source selection costly; a useful future direction is to identify statistics that predict downstream efficacy without running the full pipeline.
Second, our experiments stop at 1B parameters and 25K PT steps under compute constraints (Appendix~\ref{app:compute} reports wall-clock costs); establishing scaling laws for PPT-induced robustness is a natural follow-up. Third, we restrict the generator to RNNs as a representative instance of recurrent structure to attribute the observed effect to generic sequential structure rather than architecture-specific design; extending to LSTMs, state-space models, and other sequence architectures is a clear next direction.

%% ============================================================
\section{Conclusion}
\label{sec:conclusion}
%% ============================================================
We demonstrate that synthetic PPT improves robustness to noisy PT data without modifying the PT recipe. We evaluate this effect under controlled corruption, multiple noise types, real web noise, and model scales up to 1B parameters. Across these settings, a lightweight RNN-based PPT stage consistently yields the strongest gains, with benefits that grow as pre-training becomes noisier.

Our experiments point to two design principles. First, the synthetic source should be learnable but not trivial: overly complex generators transfer poorly, while unigram, bigram, and trigram sources are too weak to carry meaningful dependencies. Second, the source should be less biased: increasing the number of RNN generators and broadening the vocabulary both improve robustness. 
Our attention analysis offers one explanation. PPT does not reduce the model's attention to noise before PT; rather, RNN-PPT initially increases attention among corrupted tokens. Its advantage appears during noisy PT: RNN-PPT guides training away from noise self-modeling, whereas models trained without PPT increasingly model the noise itself.

%% ============================================================
%% References
%% ============================================================

\bibliographystyle{plainnat}
\bibliography{references}

%% ============================================================
%% APPENDIX
%% ============================================================

\appendix

\section{Additional Experimental Details}
\label{app:details}

Table~\ref{tab:setup} summarizes the main experimental configurations used throughout the paper.

\begin{table}[ht]
\centering
\small
\begin{tabular}{lcc}
\toprule
Setting & Main & Scale-up \\
\midrule
Model size & 160M & 1B \\
PPT steps & 500 & 1000 \\
PT steps & 10k & 25k \\
Batch size & 16 & 16 \\
Gradient accumulation & 2 & 2 \\
Effective batch size & 32 & 32 \\
Sequence length & 2048 & 2048 \\
Learning rate & $7\times10^{-4}$ & $5\times10^{-4}$ \\
LR schedule & Cosine w/ warmup & Cosine w/ warmup \\
Warmup steps & 500 & 500 \\
Weight decay & 0.1 & 0.1 \\
Gradient clipping & 1.0 & 1.0 \\
Optimizer & AdamW & AdamW \\
$\beta_1,\beta_2$ & 0.9, 0.999 & 0.9, 0.999 \\
$\epsilon$ & $10^{-6}$ & $10^{-6}$ \\
Mixed precision & bf16 & bf16 \\
\bottomrule
\end{tabular}
\caption{Summary of the main training configurations.}
\label{tab:setup}
\end{table}

We use 1000 PPT steps for these 1B runs, based on a small pilot sweep at this scale; gains beyond 1000 were marginal. All experiments reported here were run with Hugging Face Transformers 5.5.0 and Datasets 4.8.4.

Unless explicitly noted otherwise, both PPT and PT runs use \texttt{packing\_mode=pack}. Therefore, the number of tokens used for the same number of steps is effectively the same.

\paragraph{Controlled corruption variants.} For token permutation, each 2048-token sequence is partitioned into non-overlapping windows of size 32. We randomly select windows until the number of covered tokens reaches the target corruption rate, and independently permute tokens within each selected window. For span corruption, we repeatedly sample span lengths uniformly from integers 5--20 and random start positions until the union of selected span positions reaches the target corruption rate. Tokens in the selected positions are replaced with tokens sampled uniformly from the vocabulary.

\paragraph{Block training and LR schedule.} We adopt a two-stage block-training protocol in which PPT and PT are two independent optimization runs, each with its own cosine-with-warmup schedule and its own 500-step linear warmup (see Table~\ref{tab:setup}). 

Our PPT stage is lightweight: at 1B scale it occupies only $S_{\text{ppt}}=1000$ steps against a PT budget of $S_{\text{base}}=25\mathrm{K}$ steps, i.e.\ roughly $4\%$ of the baseline's PT budget (and $5\%$ at 160M, where $S_{\text{base}}=10\mathrm{K}$). Because the PPT cost is this small, whether one folds the PPT steps into the token-saving denominator has only a minor effect on the headline number. For consistency with prior works, we follow their convention and report PT-token savings rather than total-token savings. Concretely, if the baseline uses $S_{\text{base}}$ PT steps and a PPT method reaches the baseline's final loss after $S_{\text{match}}$ PT steps following a fixed $S_{\text{ppt}}$-step PPT stage, then we report $1-S_{\text{match}}/S_{\text{base}}$.

\section{Downstream Language Evaluation}
\label{app:downstream_eval}

Table~\ref{tab:lambada_ppl_160m} and Table~\ref{tab:lambada_ppl_1b} report LAMBADA perplexity for the main comparison methods. We focus on perplexity because it provides a more fine-grained language-modeling signal. All reported entries are means and standard deviations over three seeds. Lower is better. For reference, these perplexities are far from random prediction: with a vocabulary of roughly 50k tokens, a uniform-random predictor would have perplexity on the order of 50,000, whereas the values reported here are in the 76--797 range.

\begin{table}[ht]
\centering
\small
\setlength{\tabcolsep}{5pt}
\begin{tabular}{lcccc}
\toprule
Method & $0\%$ & $10\%$ & $30\%$ & $50\%$ \\
\midrule
Baseline & $404.8\pm45.9$ & $404.9\pm42.6$ & $489.0\pm32.1$ & $796.7\pm72.1$ \\
Dyck PPT & $358.1\pm40.6$ & $411.2\pm92.0$ & $511.7\pm52.7$ & $690.2\pm130.2$ \\
RNN PPT  & $\mathbf{337.6\pm39.0}$ & $\mathbf{361.2\pm75.8}$ & $\mathbf{457.0\pm63.4}$ & $\mathbf{589.9\pm8.4}$ \\
\bottomrule
\end{tabular}
\caption{LAMBADA perplexity for the main comparison methods at 160M scale.}
\label{tab:lambada_ppl_160m}
\end{table}

\begin{table}[ht]
\centering
\small
\setlength{\tabcolsep}{5pt}
\begin{tabular}{lcccc}
\toprule
Method & $0\%$ & $5\%$ & $10\%$ & $15\%$ \\
\midrule
Baseline & $113.7\pm6.4$ & $130.0\pm3.9$ & $141.8\pm7.1$ & $159.7\pm5.1$ \\
C4 PPT   & $97.5\pm9.9$ & $111.6\pm9.7$ & $122.4\pm14.3$ & $159.9\pm18.6$ \\
Dyck PPT & $84.6\pm6.1$ & $91.4\pm5.2$ & $95.7\pm8.3$ & $107.4\pm7.2$ \\
RNN PPT  & $\mathbf{76.5\pm0.6}$ & $\mathbf{83.1\pm3.4}$ & $\mathbf{86.6\pm1.6}$ & $\mathbf{94.2\pm4.1}$ \\
\bottomrule
\end{tabular}
\caption{LAMBADA perplexity for the main comparison methods at 1B scale.}
\label{tab:lambada_ppl_1b}
\end{table}
\begin{table}[ht]
\centering
\footnotesize
\setlength{\tabcolsep}{5pt}
\begin{tabular}{lcccc}
\toprule
Method & $0\%$ & $10\%$ & $30\%$ & $50\%$ \\
\midrule
Baseline & $0.605\pm0.007$ & $0.599\pm0.005$ & $0.596\pm0.004$ & $0.585\pm0.007$ \\
Dyck PPT & $0.607\pm0.007$ & $0.600\pm0.003$ & $0.596\pm0.005$ & $0.587\pm0.007$ \\
RNN PPT  & $\mathbf{0.608\pm0.006}$ & $\mathbf{0.605\pm0.002}$ & $\mathbf{0.597\pm0.007}$ & $\mathbf{0.592\pm0.005}$ \\
\bottomrule
\end{tabular}
\caption{PIQA zero-shot normalized accuracy at 160M scale. Values are mean $\pm$ standard deviation over three seeds. Random chance is $0.5$.}
\label{tab:piqa_160m}
\end{table}

\begin{table}[ht]
\centering
\footnotesize
\setlength{\tabcolsep}{5pt}
\begin{tabular}{lcccc}
\toprule
Method & $0\%$ & $5\%$ & $10\%$ & $15\%$ \\
\midrule
Baseline & $0.627\pm0.005$ & $0.626\pm0.002$ & $0.617\pm0.011$ & $0.616\pm0.007$ \\
C4 PPT   & $0.640\pm0.006$ & $0.631\pm0.008$ & $0.629\pm0.007$ & $0.623\pm0.002$ \\
Dyck PPT & $0.639\pm0.008$ & $0.634\pm0.004$ & $0.632\pm0.002$ & $0.632\pm0.009$ \\
RNN PPT  & $\mathbf{0.642\pm0.003}$ & $\mathbf{0.641\pm0.004}$ & $\mathbf{0.636\pm0.001}$ & $\mathbf{0.632\pm0.003}$ \\
\bottomrule
\end{tabular}
\caption{PIQA zero-shot normalized accuracy at 1B scale. Values are mean $\pm$ standard deviation over three seeds. Random chance is $0.5$.}
\label{tab:piqa_1b}
\end{table}

To complement LAMBADA with a standard zero-shot downstream benchmark, we evaluate on PIQA~\citep{bisk2019piqareasoningphysicalcommonsense}, a multiple-choice task in which the model scores two candidate completions of a prompt and the higher-likelihood option is selected (random chance is $0.5$). Table~\ref{tab:piqa_160m} and Table~\ref{tab:piqa_1b} report PIQA normalized accuracy. We place these results in the appendix for the same reason as LAMBADA: they serve as supplementary evidence rather than the primary basis of our claim. Higher is better.

\section{C4 Pre-Pre-Training Across Scales}
\label{app:c4ppt_scales}

In the main text we only include the C4-PPT control at 1B scale (Section~\ref{sec:scale}). For completeness, Table~\ref{tab:c4ppt_160m} extends the 160M controlled-noise comparison in Figure~\ref{fig:main_curves} to include both Random-PPT and C4-PPT alongside the Baseline, Dyck-PPT, and RNN-PPT methods. All entries are mean $\pm$ standard deviation of final C4 validation loss over three seeds. Lower is better.

\begin{table}[ht]
\centering
\small
\setlength{\tabcolsep}{5pt}
\begin{tabular}{lcccc}
\toprule
Method & $0\%$ & $10\%$ & $30\%$ & $50\%$ \\
\midrule
Baseline & $3.627\pm0.001$ & $3.659\pm0.003$ & $3.719\pm0.003$ & $3.818\pm0.017$ \\
Random PPT & $3.621\pm0.002$ & $3.654\pm0.004$ & $3.710\pm0.006$ & $3.792\pm0.005$ \\
Dyck PPT & $3.619\pm0.000$ & $3.645\pm0.001$ & $3.703\pm0.003$ & $3.798\pm0.006$ \\
C4 PPT   & $3.613\pm0.004$ & $3.635\pm0.003$ & $3.703\pm0.005$ & $3.786\pm0.007$ \\
RNN PPT  & $\mathbf{3.603\pm0.002}$ & $\mathbf{3.628\pm0.005}$ & $\mathbf{3.681\pm0.004}$ & $\mathbf{3.761\pm0.003}$ \\
\bottomrule
\end{tabular}
\caption{Final C4 validation loss at 160M scale with Random-PPT and C4-PPT included. Values are mean $\pm$ standard deviation across three seeds.}
\label{tab:c4ppt_160m}
\end{table}

RNN-PPT remains the strongest method at every tested noise level, improving validation loss over C4-PPT by $0.010$ at $0\%$ noise and by $0.025$ at $50\%$ noise. Random-PPT, by contrast, stays close to the baseline throughout, which is consistent with the main-text interpretation that warm-up alone is not sufficient. 

\section{Effect of Extended PT Budget}
\label{app:longpt}

We extend the 160M PT budget from 10k to 20k steps at noise rates $\{0\%, 10\%, 30\%, 50\%\}$ and compare RNN-PPT with Baseline models trained fully from scratch. Table~\ref{tab:longpt} reports the final C4 validation loss at 20k PT steps across three seeds. RNN-PPT remains better than the baseline at every tested noise level.

\begin{table}[ht]
\centering
\small
\setlength{\tabcolsep}{5pt}
\begin{tabular}{ccccc}
\toprule
Noise & PT steps & Baseline & RNN-PPT & $\Delta$ (Baseline $-$ RNN) \\
\midrule
$0\%$   & 20k & $3.506 \pm 0.001$ & $3.485 \pm 0.001$ & $+0.021 \pm 0.001$ \\
$10\%$  & 20k & $3.529 \pm 0.002$ & $3.508 \pm 0.003$ & $+0.021 \pm 0.004$ \\
$30\%$  & 20k & $3.582 \pm 0.002$ & $3.554 \pm 0.001$ & $+0.028 \pm 0.003$ \\
$50\%$  & 20k & $3.656 \pm 0.007$ & $3.625 \pm 0.002$ & $+0.031 \pm 0.005$ \\
\bottomrule
\end{tabular}
\caption{Final C4 validation loss at 160M scale with Baseline and RNN-PPT, reported at PT step 20k as mean $\pm$ standard deviation across three seeds. RNN-PPT remains better than the baseline across all tested noise levels.}
\label{tab:longpt}
\end{table}

\section{Cross-Noise RNN Design Ablations}
\label{app:rnn_ablation_noise}

For readability, Figure~\ref{fig:rnn_ablation} in the main text plots the $0\%$ and $30\%$ PT settings. Table~\ref{tab:rnn_generators_extra_noise}, Table~\ref{tab:rnn_hidden_extra_noise}, and Table~\ref{tab:rnn_vocab_extra_noise} report the complementary $10\%$ and $50\%$ noise levels for the same three ablation families, so together they cover the full $\{0,10,30,50\}\%$ sweep. Lower is better. The full sweep supports the interpretation in the main text. 

\begin{table*}[t]
\centering
\small
\setlength{\tabcolsep}{5pt}
\resizebox{\textwidth}{!}{%
\begin{tabular}{lcccccccc}
\toprule
PT noise & Baseline & RNN-$m{=}1$ & RNN-$m{=}10$ & RNN-$m{=}100$ & RNN-$m{=}500$ & RNN-$m{=}1000$ & RNN-$m{=}2000$ & RNN-$m{=}3000$ \\
\midrule
$10\%$ & $3.6581$ & $3.6607$ & $3.6449$ & $3.6310$ & $3.6302$ & $\mathbf{3.6225}$ & $3.6254$ & $3.6370$ \\
$50\%$ & $3.8250$ & $3.8008$ & $3.8143$ & $3.7766$ & $3.7763$ & $\mathbf{3.7700}$ & $3.7789$ & $3.7917$ \\
\bottomrule
\end{tabular}%
}
\caption{Generator-count ablation at the complementary $10\%$ and $50\%$ PT noise levels. Together with the $0\%$ and $30\%$ curves in Figure~\ref{fig:rnn_ablation}, these results support the low-bias principle: performance improves once the ensemble reaches roughly $100$ generators and is strongest around $1000$--$3000$.}
\label{tab:rnn_generators_extra_noise}
\end{table*}

\begin{table*}[t]
\centering
\small
\setlength{\tabcolsep}{5pt}
\resizebox{\textwidth}{!}{%
\begin{tabular}{lccccccccc}
\toprule
PT noise & Baseline & RNN-$h{=}8$ & RNN-$h{=}16$ & RNN-$h{=}32$ & RNN-$h{=}64$ & RNN-$h{=}128$ & RNN-$h{=}256$ & RNN-$h{=}512$ & RNN-$h{=}1024$ \\
\midrule
$10\%$ & $3.6577$ & $3.6290$ & $3.6282$ & $3.6302$ & $\mathbf{3.6255}$ & $3.6300$ & $3.6301$ & $3.6562$ & $3.6400$ \\
$50\%$ & $3.8367$ & $3.7838$ & $\mathbf{3.7652}$ & $3.7769$ & $3.7750$ & $3.7753$ & $3.7860$ & $3.8140$ & $3.8143$ \\
\bottomrule
\end{tabular}%
}
\caption{Hidden-size ablation at the complementary $10\%$ and $50\%$ PT noise levels. The full sweep preserves the same broader pattern: generators in a learnable complexity range ($h\in\{16,32,64\}$) transfer best, while large generators ($h\in\{512,1024\}$) lose most of the gain.}
\label{tab:rnn_hidden_extra_noise}
\end{table*}

\begin{table*}[t]
\centering
\small
\setlength{\tabcolsep}{5pt}
\resizebox{0.82\textwidth}{!}{%
\begin{tabular}{lccccccc}
\toprule
PT noise & Baseline & RNN-$V{=}64$ & RNN-$V{=}256$ & RNN-$V{=}1024$ & RNN-$V{=}4096$ & RNN-$V{=}16384$ & RNN-$V{=}50304$ \\
\midrule
$10\%$ & $3.6577$ & $3.6441$ & $3.6378$ & $3.6364$ & $3.6273$ & $3.6267$ & $\mathbf{3.6248}$ \\
$50\%$ & $3.8294$ & $3.7998$ & $3.7958$ & $3.7995$ & $3.7801$ & $3.7778$ & $\mathbf{3.7692}$ \\
\bottomrule
\end{tabular}%
}
\caption{Vocabulary-size ablation at the complementary $10\%$ and $50\%$ PT noise levels. Larger vocabularies are strong overall, consistent with using broad token support to reduce source-specific bias.}
\label{tab:rnn_vocab_extra_noise}
\end{table*}

\section{Attention-Probe Protocol and Interpretation}
\label{app:attention_probe}

The attention analyses in Section~\ref{sec:attention} use a held-out probe set with independently injected corruption and do not reuse the training batches themselves. This probe uses token-level corruption: within each clean C4 validation sequence, we randomly replace the specified fraction of token positions and record those positions as the noisy-token mask. For the head-level heatmap in Figure~\ref{fig:attention_head_heatmap}, we apply a single fixed high-noise probe across all compared checkpoints so that the per-head contrasts are measured under the same evaluation condition. This probe corruption rate is $50\%$.

\section{RNN Generator Architecture}
\label{app:rnn_generator}

Our RNN-based PPT source is generated by a simple recurrent language model. For vocabulary size $V$ and hidden size $H$, each generator maintains a hidden state $h_t\in\mathbb{R}^H$ and updates it according to
\begin{equation}
h_t = A e_{x_{t-1}} + W h_{t-1} + b,
\qquad
\ell_t = C h_t + d,
\end{equation}
where $e_{x_{t-1}}\in\mathbb{R}^V$ is the one-hot vector of the previous token, $A\in\mathbb{R}^{H\times V}$ is the input projection, $W\in\mathbb{R}^{H\times H}$ is the recurrent matrix, and $C\in\mathbb{R}^{V\times H}$ is the output projection. The next-token distribution is $p_t=\mathrm{softmax}(\ell_t / \tau)$, and the next token is sampled autoregressively as $x_t\sim\mathrm{Categorical}(p_t)$, with temperature $\tau=1$ in the default setup. The hidden state is initialized to zero, and the first token is sampled uniformly from the vocabulary. We do not sample variable-length sequences: each raw RNN sequence is generated at length 2048 before training-time packing. This choice keeps the synthetic source simple and stable while still producing temporal dependencies through recurrence, autoregressive feedback, and token sampling.

All RNN parameters are sampled once at random and then kept fixed; the generators themselves are never trained. In the implementation, entries of $A$ are sampled i.i.d.\ from a zero-mean Gaussian with standard deviation $1/\sqrt{V}$, while entries of $W$ and $C$ are sampled i.i.d.\ with standard deviation $1/\sqrt{H}$. The bias terms $b$ and $d$ are initialized to zero. To keep the dynamics stable while preserving long-range mixing, we rescale $W$ so that its estimated spectral radius, computed by power iteration, is $0.9$.

\section{Compute Budget and Runtime}
\label{app:compute}

To contextualize our choice of scale, Table~\ref{tab:compute} summarizes representative wall-clock runtimes measured on a single NVIDIA H200 GPU. The synthetic PPT stage is lightweight: it typically takes only about 5-10 minutes for 500 steps. 

The main PT runs are much more expensive. At 160M scale, a standard 10K-step PT run takes about 1 hour. At 1B scale, the 25K-step PT runs used in Section~\ref{sec:scale} take about 22 hours each. Taking into account the experiments across different random seeds and different noise levels, the aggregate compute budget is substantial.

All synthetic data preparation for Dyck, Random, and RNN PPT sources is performed on CPU and completes within an hour on our 96-core server. These data can be reused across experiments and do not dominate the end-to-end training workflow.

\begin{table}[ht]
\centering
\small
\begin{tabular}{lccc}
\toprule
Configuration & Stage & Budget & Approx. wall-clock \\
\midrule
160M main setting & synthetic PPT & 500 steps & $\sim$5 min \\
160M main setting & PT & 10K steps & $\sim$1 hour \\
1B scale-up setting & synthetic PPT & 1000 steps & $\sim$50 min \\
1B scale-up setting & PT & 25K steps & $\sim$22 hours \\
\bottomrule
\end{tabular}
\caption{Representative runtime costs on our hardware.}
\label{tab:compute}
\end{table}

\section{FineWeb Quality-Score Partition}
\label{sec:fineweb_partition}

For the natural-noise experiment, we begin with a tokenized 3M-document FineWeb subset. We score each document with a frozen external causal language model, \texttt{openlm-research/open\_llama\_3b\_v2}~\citep{openlm2023openllama}, and use the model's mean next-token cross-entropy over the document as a scalar quality score. We choose this reference model for two reasons. First, reference-model-based filtering is a strong baseline for data pruning, and recent evidence shows that even relatively small reference models can identify high-quality subsets that improve downstream pre-training performance~\citep{ankner2024perplexed}. Second, OpenLLaMA-3B-v2 is a different model family from the Pythia architectures used in our main experiments and was trained on a broad public mixture rather than on C4 alone, including Falcon RefinedWeb, StarCoderData, and the Wikipedia/arXiv/books/StackExchange portions of RedPajama, which helps decouple the partition from our downstream training setup.

We sort all documents by this score and retain only the two extremes of the ranking used in the main paper. The high-quality subset is drawn from the bottom third of the score distribution, and the low-quality subset from the top third. The corresponding cutoffs are the 33rd and 67th percentiles of the full distribution. This means the split is fully determined by a fixed external scorer and percentile thresholds, with no manual document selection. To keep the PT budget comparable across subsets, we traverse documents within each retained subset in score order and accumulate them until reaching roughly 660M tokens.

This procedure yields two natural PT conditions that differ in document quality while keeping the training recipe otherwise fixed. The main text reports only these two retained subsets because they provide the clearest high-quality versus low-quality comparison.

The reference-model scoring step is a one-time preprocessing cost. On a single NVIDIA H200 GPU in bfloat16, scoring 660M tokens with the 3B reference model takes about 4 hours.

\section{Representative Cases of Naturally Clean and Noisy Data}
\label{app:fineweb_cases}

Table~\ref{tab:fineweb_cases} shows representative cases from the actual FineWeb subsets used in Section~\ref{sec:fineweb}. We include two examples from the retained high-quality subset and two from the retained low-quality subset. The cleaner subset more often contains coherent article-style prose, while the noisier subset more often contains templated listings, repetitive boilerplate, and ad-like fragments.

In pre-training pipelines, these templated listings, directory dumps, SEO boilerplate, and ad-like fragments are precisely the kinds of content that are regarded as noise and routinely filtered out in practice~\citep{penedo2024fineweb}. Prior empirical studies likewise show that noisy or low-quality data can weaken pre-training outcomes, and that filtering decisions involve tradeoffs between quality and coverage~\citep{zhang2026empiricalstudynoisydata,longpre2023pretrainersguidetrainingdata}. Their defining limitation is therefore not the absence of surface structure, but the scarcity of transferable linguistic signal: although they can contain repeated templates and other local surface regularities, they contribute little generalizable information. From this perspective, this partition captures a realistic and practically relevant form of noise that models encounter at web scale.

\begin{table*}[t]
\centering
\small
\setlength{\tabcolsep}{5pt}
\begin{tabular}{p{0.16\textwidth}p{0.16\textwidth}p{0.60\textwidth}}
\toprule
Subset & Example type & Case \\
\midrule
Cleaner & Entertainment article & If you haven't seen the new trailer for The Lego Batman Movie, then you are missing out on one of the best laughs you will have all day. Not only is Lego Batman too good to be a side character in The Lego Movie, but Lego Batman rightfully deserves his own film. \ldots{} \\
Cleaner & News report & CONWAY, Ark. (AP) -- LaQuentin Miles scored 20 points, including 10 of Central Arkansas' final 11 points, and the Bears held on to beat Southern Illinois-Edwardsville 80-78 on Saturday. Jarvis Garner added 19 points and Robert Crawford scored 18 for Central Arkansas. \ldots{} \\
Noisier & Directory / ad mix & IFSC code: ICIC0002980 and MICR code: 342229203; ICICI BANK DANWARA address: Icici Bank Ltd., Village Danwara, Pin - 342037, Tehsil Baori, Dist. Jodhpur, Rajasthan; Branch code is 002980. Credit Score of 750 = Easy approval. \ldots{} \\
Noisier & Listing / boilerplate & Use Cvent to book the G Casino Sheffield in Sheffield, England for your event and get a great rate. Find upcoming events at Grosvenor G Casino Sheffield in Sheffield. Full event details plus travel info, opening times + venue info. This table is no longer available, followed by unrelated multilingual boilerplate. \ldots{} \\
\bottomrule
\end{tabular}
\caption{Representative cases from the FineWeb subsets used in the natural-noise experiment. Snippets are abbreviated for readability.}
\label{tab:fineweb_cases}
\end{table*}

\section{Unigram, Bigram, and Trigram Metamer Construction Details}
\label{app:metamer}

This appendix provides the construction details for the unigram, bigram, and trigram metamer diagnostic reported in Section~\ref{sec:metamer} and Table~\ref{tab:metamer}.

\paragraph{RNN-subset.} The main-paper RNN-PPT source is generated by an ensemble of 1000 randomly initialized RNNs. Fitting a full-vocabulary ($V=50{,}304$) trigram count table over the entire source is storage-prohibitive, so we first construct an ``RNN-subset'' by uniformly sampling sequences from the full 1000-generator source. The subset preserves the original generator-level diversity. Its size is set so that a 500-step PPT run at the main-paper batch size and sequence length (effective batch 32, sequence length 2048) consumes strictly less than one epoch of the subset; consequently, training never revisits a token.

\paragraph{Metamer fitting and sampling.} For each order $n\in\{1,2,3\}$, we fit the corresponding model on exactly the same RNN-subset, so that the metamer statistics come from the same token distribution as the reference condition. We then sample new token sequences autoregressively from the fitted model, drawing a fresh stream whose total token count matches the PPT token budget used by every other PPT condition. Unigram, bigram, trigram, and RNN-subset conditions therefore all see the same number of PPT tokens with no token repetition during training.
For trigram sampling, each sequence is initialized from an observed start bigram. At generation time, if the current bigram context has no observed trigram continuation, we back off to the empirical bigram model conditioned on the last token; if that context also has no observed continuation, we sample from the empirical unigram distribution. Bigram sampling uses the analogous unigram back-off.

%%%%%%%%%%%%%%%%%%%%%%%%%%%%%%%%%%%%%%%%%%%%%%%%%%%%%%%%%%%%

\newpage
\input{checklist.tex}

\end{document}

%% file: checklist.tex
\section*{NeurIPS Paper Checklist}

\begin{enumerate}

\item {\bf Claims}
    \item[] Question: Do the main claims made in the abstract and introduction accurately reflect the paper's contributions and scope?
    \item[] Answer: \answerYes{}
    \item[] Justification: The claims are stated in the abstract and Section~\ref{sec:intro}, and are supported by the experiments in Sections~\ref{sec:main_results}--\ref{sec:scale}.

\item {\bf Limitations}
    \item[] Question: Does the paper discuss the limitations of the work performed by the authors?
    \item[] Answer: \answerYes{}
    \item[] Justification: Section~\ref{sec:discussion} discusses limitations such as model scale, PT budget, source selection cost, and the restriction to RNN generators.

\item {\bf Theory assumptions and proofs}
    \item[] Question: For each theoretical result, does the paper provide the full set of assumptions and a complete (and correct) proof?
    \item[] Answer: \answerNA{}
    \item[] Justification: The paper does not present theoretical results requiring formal proofs.

    \item {\bf Experimental result reproducibility}
    \item[] Question: Does the paper fully disclose all the information needed to reproduce the main experimental results of the paper to the extent that it affects the main claims and/or conclusions of the paper (regardless of whether the code and data are provided or not)?
    \item[] Answer: \answerYes{}
    \item[] Justification: Section~\ref{sec:setup} and Appendix~\ref{app:details} provide the main datasets, model settings, hyperparameters, and evaluation details.

\item {\bf Open access to data and code}
    \item[] Question: Does the paper provide open access to the data and code, with sufficient instructions to faithfully reproduce the main experimental results, as described in supplemental material?
    \item[] Answer: \answerYes{}
    \item[] Justification: The datasets and models used are public, and we can provide the code link at any time. For the review version, we did not include the link to keep the submission anonymous.

\item {\bf Experimental setting/details}
    \item[] Question: Does the paper specify all the training and test details (e.g., data splits, hyperparameters, how they were chosen, type of optimizer) necessary to understand the results?
    \item[] Answer: \answerYes{}
    \item[] Justification: Section~\ref{sec:setup} and Appendix~\ref{app:details} give the dataset, model, optimizer, hyperparameter, and evaluation details.

\item {\bf Experiment statistical significance}
    \item[] Question: Does the paper report error bars suitably and correctly defined or other appropriate information about the statistical significance of the experiments?
    \item[] Answer: \answerYes{}
    \item[] Justification: Results are averaged over three seeds, with standard deviations or error bars reported in the main figures and appendix tables.

\item {\bf Experiments compute resources}
    \item[] Question: For each experiment, does the paper provide sufficient information on the computer resources (type of compute workers, memory, time of execution) needed to reproduce the experiments?
    \item[] Answer: \answerYes{}
    \item[] Justification: Appendix~\ref{app:compute} reports representative runtimes on a single NVIDIA H200 GPU, and Appendix~\ref{app:details} gives the training budgets.
    
\item {\bf Code of ethics}
    \item[] Question: Does the research conducted in the paper conform, in every respect, with the NeurIPS Code of Ethics \url{https://neurips.cc/public/EthicsGuidelines}?
    \item[] Answer: \answerYes{}
    \item[] Justification: The work uses public datasets/models and synthetic generators, and does not involve human subjects.

\item {\bf Broader impacts}
    \item[] Question: Does the paper discuss both potential positive societal impacts and negative societal impacts of the work performed?
    \item[] Answer: \answerNo{}
    \item[] Justification: The paper is mainly foundational and does not include a dedicated discussion of both positive and negative societal impacts.
    
\item {\bf Safeguards}
    \item[] Question: Does the paper describe safeguards that have been put in place for responsible release of data or models that have a high risk for misuse (e.g., pre-trained language models, image generators, or scraped datasets)?
    \item[] Answer: \answerNA{}
    \item[] Justification: The submission does not release a pretrained model, image generator, scraped dataset, or other high-risk artifact.

\item {\bf Licenses for existing assets}
    \item[] Question: Are the creators or original owners of assets (e.g., code, data, models), used in the paper, properly credited and are the license and terms of use explicitly mentioned and properly respected?
    \item[] Answer: \answerYes{}
    \item[] Justification: All the assets in this work, either are original or publicly available (e.g., data, models).

\item {\bf New assets}
    \item[] Question: Are new assets introduced in the paper well documented and is the documentation provided alongside the assets?
    \item[] Answer: \answerNA{}
    \item[] Justification: The paper does not release new datasets, code packages, model checkpoints, or other assets.

\item {\bf Crowdsourcing and research with human subjects}
    \item[] Question: For crowdsourcing experiments and research with human subjects, does the paper include the full text of instructions given to participants and screenshots, if applicable, as well as details about compensation (if any)? 
    \item[] Answer: \answerNA{}
    \item[] Justification: The research does not involve crowdsourcing experiments or human subjects.

\item {\bf Institutional review board (IRB) approvals or equivalent for research with human subjects}
    \item[] Question: Does the paper describe potential risks incurred by study participants, whether such risks were disclosed to the subjects, and whether Institutional Review Board (IRB) approvals (or an equivalent approval/review based on the requirements of your country or institution) were obtained?
    \item[] Answer: \answerNA{}
    \item[] Justification: The research does not involve crowdsourcing or human subjects.

\item {\bf Declaration of LLM usage}
    \item[] Question: Does the paper describe the usage of LLMs if it is an important, original, or non-standard component of the core methods in this research? Note that if the LLM is used only for writing, editing, or formatting purposes and does \emph{not} impact the core methodology, scientific rigor, or originality of the research, declaration is not required.
    %this research?
    \item[] Answer: \answerYes{}
    \item[] Justification: We pre-train Pythia-architecture LLMs (160M and 1B parameters) as the subjects of all experiments (Section~\ref{sec:exp}), and use a frozen external LLM, \texttt{openlm-research/open\_llama\_3b\_v2}, as a reference model for the FineWeb quality-score partition (Appendix~\ref{sec:fineweb_partition}).

\end{enumerate}

%% file: references.bib
@misc{hu2025prepretraining,
  title         = {Between Circuits and Chomsky: Pre-pretraining on Formal Languages Imparts Linguistic Biases},
  author        = {Michael Y. Hu and Jackson Petty and Chuan Shi and William Merrill and Tal Linzen},
  year          = {2025},
  eprint        = {2502.19249},
  archiveprefix = {arXiv},
  primaryclass  = {cs.CL},
  url           = {https://arxiv.org/abs/2502.19249}
}

@misc{papadimitriou2020learning,
  title         = {Learning Music Helps You Read: Using Transfer to Study Linguistic Structure in Language Models},
  author        = {Isabel Papadimitriou and Dan Jurafsky},
  year          = {2020},
  eprint        = {2004.14601},
  archiveprefix = {arXiv},
  primaryclass  = {cs.CL},
  url           = {https://arxiv.org/abs/2004.14601}
}

@misc{papadimitriou2023injecting,
  title         = {Injecting structural hints: Using language models to study inductive biases in language learning},
  author        = {Isabel Papadimitriou and Dan Jurafsky},
  year          = {2023},
  eprint        = {2304.13060},
  archiveprefix = {arXiv},
  primaryclass  = {cs.CL},
  url           = {https://arxiv.org/abs/2304.13060}
}

@misc{paperno2016lambada,
      title={The LAMBADA dataset: Word prediction requiring a broad discourse context}, 
      author={Denis Paperno and Germán Kruszewski and Angeliki Lazaridou and Quan Ngoc Pham and Raffaella Bernardi and Sandro Pezzelle and Marco Baroni and Gemma Boleda and Raquel Fernández},
      year={2016},
      eprint={1606.06031},
      archivePrefix={arXiv},
      primaryClass={cs.CL},
      url={https://arxiv.org/abs/1606.06031}, 
}

@misc{biderman2023pythia,
      title={Pythia: A Suite for Analyzing Large Language Models Across Training and Scaling}, 
      author={Stella Biderman and Hailey Schoelkopf and Quentin Anthony and Herbie Bradley and Kyle O'Brien and Eric Hallahan and Mohammad Aflah Khan and Shivanshu Purohit and USVSN Sai Prashanth and Edward Raff and Aviya Skowron and Lintang Sutawika and Oskar van der Wal},
      year={2023},
      eprint={2304.01373},
      archivePrefix={arXiv},
      primaryClass={cs.CL},
      url={https://arxiv.org/abs/2304.01373}, 
}

@article{raffel2020exploring,
  title={Exploring the limits of transfer learning with a unified text-to-text transformer},
  author={Raffel, Colin and Shazeer, Noam and Roberts, Adam and Lee, Katherine and Narang, Sharan and Matena, Michael and Zhou, Yanqi and Li, Wei and Liu, Peter J},
  journal={Journal of machine learning research},
  volume={21},
  number={140},
  pages={1--67},
  year={2020}
}

@misc{gunasekar2023textbooks,
      title={Textbooks Are All You Need}, 
      author={Suriya Gunasekar and Yi Zhang and Jyoti Aneja and Caio César Teodoro Mendes and Allie Del Giorno and Sivakanth Gopi and Mojan Javaheripi and Piero Kauffmann and Gustavo de Rosa and Olli Saarikivi and Adil Salim and Shital Shah and Harkirat Singh Behl and Xin Wang and Sébastien Bubeck and Ronen Eldan and Adam Tauman Kalai and Yin Tat Lee and Yuanzhi Li},
      year={2023},
      eprint={2306.11644},
      archivePrefix={arXiv},
      primaryClass={cs.CL},
      url={https://arxiv.org/abs/2306.11644}, 
}

@misc{penedo2024fineweb,
      title={The FineWeb Datasets: Decanting the Web for the Finest Text Data at Scale}, 
      author={Guilherme Penedo and Hynek Kydlíček and Loubna Ben allal and Anton Lozhkov and Margaret Mitchell and Colin Raffel and Leandro Von Werra and Thomas Wolf},
      year={2024},
      eprint={2406.17557},
      archivePrefix={arXiv},
      primaryClass={cs.CL},
      url={https://arxiv.org/abs/2406.17557}, 
}

@misc{ankner2024perplexed,
      title={Perplexed by Perplexity: Perplexity-Based Data Pruning With Small Reference Models}, 
      author={Zachary Ankner and Cody Blakeney and Kartik Sreenivasan and Max Marion and Matthew L. Leavitt and Mansheej Paul},
      year={2024},
      eprint={2405.20541},
      archivePrefix={arXiv},
      primaryClass={cs.LG},
      url={https://arxiv.org/abs/2405.20541}, 
}

@software{openlm2023openllama,
  author = {Geng, Xinyang and Liu, Hao},
  title = {OpenLLaMA: An Open Reproduction of LLaMA},
  month = May,
  year = 2023,
  url = {https://github.com/openlm-research/open_llama}
}

@misc{rae2022gopher,
      title={Scaling Language Models: Methods, Analysis \& Insights from Training Gopher},
      author={Jack W. Rae and Sebastian Borgeaud and Trevor Cai and Katie Millican and Jordan Hoffmann and Francis Song and John Aslanides and Sarah Henderson and Roman Ring and Susannah Young and Eliza Rutherford and Tom Hennigan and Jacob Menick and Albin Cassirer and Richard Powell and George van den Driessche and Lisa Anne Hendricks and Maribeth Rauh and Po-Sen Huang and Amelia Glaese and Johannes Welbl and Sumanth Dathathri and Saffron Huang and Jonathan Uesato and John Mellor and Irina Higgins and Antonia Creswell and Nat McAleese and Amy Wu and Erich Elsen and Siddhant Jayakumar and Elena Buchatskaya and David Budden and Esme Sutherland and Karen Simonyan and Michela Paganini and Laurent Sifre and Lena Martens and Xiang Lorraine Li and Adhiguna Kuncoro and Aida Nematzadeh and Elena Gribovskaya and Domenic Donato and Angeliki Lazaridou and Arthur Mensch and Jean-Baptiste Lespiau and Maria Tsimpoukelli and Nikolai Grigorev and Doug Fritz and Thibault Sottiaux and Mantas Pajarskas and Toby Pohlen and Zhitao Gong and Daniel Toyama and Cyprien de Masson d'Autume and Yujia Li and Tayfun Terzi and Vladimir Mikulik and Igor Babuschkin and Aidan Clark and Diego de Las Casas and Aurelia Guy and Chris Jones and James Bradbury and Matthew Johnson and Blake Hechtman and Laura Weidinger and Iason Gabriel and William Isaac and Ed Lockhart and Simon Osindero and Laura Rimell and Chris Dyer and Oriol Vinyals and Kareem Ayoub and Jeff Stanway and Lorrayne Bennett and Demis Hassabis and Koray Kavukcuoglu and Geoffrey Irving},
      year={2022},
      eprint={2112.11446},
      archivePrefix={arXiv},
      primaryClass={cs.CL},
      url={https://arxiv.org/abs/2112.11446},
}

@misc{penedo2023refinedweb,
      title={The RefinedWeb Dataset for Falcon LLM: Outperforming Curated Corpora with Web Data, and Web Data Only}, 
      author={Guilherme Penedo and Quentin Malartic and Daniel Hesslow and Ruxandra Cojocaru and Alessandro Cappelli and Hamza Alobeidli and Baptiste Pannier and Ebtesam Almazrouei and Julien Launay},
      year={2023},
      eprint={2306.01116},
      archivePrefix={arXiv},
      primaryClass={cs.CL},
      url={https://arxiv.org/abs/2306.01116}, 
}

@misc{wenzek2020ccnet,
      title={CCNet: Extracting High Quality Monolingual Datasets from Web Crawl Data}, 
      author={Guillaume Wenzek and Marie-Anne Lachaux and Alexis Conneau and Vishrav Chaudhary and Francisco Guzmán and Armand Joulin and Edouard Grave},
      year={2019},
      eprint={1911.00359},
      archivePrefix={arXiv},
      primaryClass={cs.CL},
      url={https://arxiv.org/abs/1911.00359}, 
}

@misc{lee2023deduplicating,
      title={Deduplicating Training Data Makes Language Models Better}, 
      author={Katherine Lee and Daphne Ippolito and Andrew Nystrom and Chiyuan Zhang and Douglas Eck and Chris Callison-Burch and Nicholas Carlini},
      year={2022},
      eprint={2107.06499},
      archivePrefix={arXiv},
      primaryClass={cs.CL},
      url={https://arxiv.org/abs/2107.06499}, 
}

@misc{kaplan2020scalinglawsneurallanguage,
  title         = {Scaling Laws for Neural Language Models},
  author        = {Jared Kaplan and Sam McCandlish and Tom Henighan and Tom B. Brown and Benjamin Chess and Rewon Child and Scott Gray and Alec Radford and Jeffrey Wu and Dario Amodei},
  year          = {2020},
  eprint        = {2001.08361},
  archiveprefix = {arXiv},
  primaryclass  = {cs.LG},
  url           = {https://arxiv.org/abs/2001.08361}
}

@misc{bisk2019piqareasoningphysicalcommonsense,
  title         = {PIQA: Reasoning about Physical Commonsense in Natural Language},
  author        = {Yonatan Bisk and Rowan Zellers and Ronan Le Bras and Jianfeng Gao and Yejin Choi},
  year          = {2019},
  eprint        = {1911.11641},
  archiveprefix = {arXiv},
  primaryclass  = {cs.CL},
  url           = {https://arxiv.org/abs/1911.11641}
}

@misc{zhang2026empiricalstudynoisydata,
  title         = {An Empirical Study on Noisy Data and LLM Pretraining Loss Divergence},
  author        = {Qizhen Zhang and Ankush Garg and Jakob Foerster and Niladri Chatterji and Kshitiz Malik and Mike Lewis},
  year          = {2026},
  eprint        = {2602.02400},
  archiveprefix = {arXiv},
  primaryclass  = {cs.LG},
  url           = {https://arxiv.org/abs/2602.02400}
}

@misc{longpre2023pretrainersguidetrainingdata,
  title         = {A Pretrainer's Guide to Training Data: Measuring the Effects of Data Age, Domain Coverage, Quality and Toxicity},
  author        = {Shayne Longpre and Gregory Yauney and Emily Reif and Katherine Lee and Adam Roberts and Barret Zoph and Denny Zhou and Jason Wei and Kevin Robinson and David Mimno and Daphne Ippolito},
  year          = {2023},
  eprint        = {2305.13169},
  archiveprefix = {arXiv},
  primaryclass  = {cs.CL},
  url           = {https://arxiv.org/abs/2305.13169}
}

@misc{bloem2025universalpretraining,
  title         = {Universal pre-training by iterated random computation},
  author        = {Peter Bloem},
  year          = {2025},
  eprint        = {2506.20057},
  archiveprefix = {arXiv},
  primaryclass  = {cs.LG},
  url           = {https://arxiv.org/abs/2506.20057}
}

@misc{shinnick2025transformerspretrainedproceduraldata,
  title         = {Transformers Pretrained on Procedural Data Contain Modular Structures for Algorithmic Reasoning},
  author        = {Zachary Shinnick and Liangze Jiang and Hemanth Saratchandran and Anton van den Hengel and Damien Teney},
  year          = {2025},
  eprint        = {2505.22308},
  archiveprefix = {arXiv},
  primaryclass  = {cs.LG},
  url           = {https://arxiv.org/abs/2505.22308}
}

@misc{jiang2026proceduralpretrainingwarminglanguage,
  title         = {Procedural Pretraining: Warming Up Language Models with Abstract Data},
  author        = {Liangze Jiang and Zachary Shinnick and Anton van den Hengel and Hemanth Saratchandran and Damien Teney},
  year          = {2026},
  eprint        = {2601.21725},
  archiveprefix = {arXiv},
  primaryclass  = {cs.CL},
  url           = {https://arxiv.org/abs/2601.21725}
}

@misc{ru2025reallyfilterrandomnoise,
  title         = {Do we really have to filter out random noise in pre-training data for language models?},
  author        = {Jinghan Ru and Yuxin Xie and Xianwei Zhuang and Yuguo Yin and Zhihui Guo and Zhiming Liu and Qianli Ren and Yuexian Zou},
  year          = {2025},
  eprint        = {2502.06604},
  archiveprefix = {arXiv},
  primaryclass  = {cs.CL},
  url           = {https://arxiv.org/abs/2502.06604}
}

@inproceedings{siegelmann1992computational,
  title={On the computational power of neural nets},
  author={Siegelmann, Hava T and Sontag, Eduardo D},
  booktitle={Proceedings of the fifth annual workshop on Computational learning theory},
  pages={440--449},
  year={1992}
}

@article{jaeger2001echo,
  title={The “echo state” approach to analysing and training recurrent neural networks-with an erratum note},
  author={Jaeger, Herbert},
  journal={Bonn, Germany: German national research center for information technology gmd technical report},
  volume={148},
  number={34},
  pages={13},
  year={2001},
  publisher={Bonn}
}

@inproceedings{merrill2020formal,
  title={A formal hierarchy of RNN architectures},
  author={Merrill, William and Weiss, Gail and Goldberg, Yoav and Schwartz, Roy and Smith, Noah A and Yahav, Eran},
  booktitle={Proceedings of the 58th Annual Meeting of the Association for Computational Linguistics},
  pages={443--459},
  year={2020}
}

@misc{li2025datacomplmsearchgenerationtraining,
      title={{DataComp-LM}: In search of the next generation of training sets for language models}, 
      author={Jeffrey Li and Alex Fang and Georgios Smyrnis and Maor Ivgi and Matt Jordan and Samir Gadre and Hritik Bansal and Etash Guha and Sedrick Keh and Kushal Arora and Saurabh Garg and Rui Xin and Niklas Muennighoff and Reinhard Heckel and Jean Mercat and Mayee Chen and Suchin Gururangan and Mitchell Wortsman and Alon Albalak and Yonatan Bitton and Marianna Nezhurina and Amro Abbas and Cheng-Yu Hsieh and Dhruba Ghosh and Josh Gardner and Maciej Kilian and Hanlin Zhang and Rulin Shao and Sarah Pratt and Sunny Sanyal and Gabriel Ilharco and Giannis Daras and Kalyani Marathe and Aaron Gokaslan and Jieyu Zhang and Khyathi Chandu and Thao Nguyen and Igor Vasiljevic and Sham Kakade and Shuran Song and Sujay Sanghavi and Fartash Faghri and Sewoong Oh and Luke Zettlemoyer and Kyle Lo and Alaaeldin El-Nouby and Hadi Pouransari and Alexander Toshev and Stephanie Wang and Dirk Groeneveld and Luca Soldaini and Pang Wei Koh and Jenia Jitsev and Thomas Kollar and Alexandros G. Dimakis and Yair Carmon and Achal Dave and Ludwig Schmidt and Vaishaal Shankar},
      year={2025},
      eprint={2406.11794},
      archivePrefix={arXiv},
      primaryClass={cs.LG},
      url={https://arxiv.org/abs/2406.11794}, 
}

@misc{soldaini2024dolmaopencorpustrillion,
      title={{Dolma}: an Open Corpus of Three Trillion Tokens for Language Model Pretraining Research}, 
      author={Luca Soldaini and Rodney Kinney and Akshita Bhagia and Dustin Schwenk and David Atkinson and Russell Authur and Ben Bogin and Khyathi Chandu and Jennifer Dumas and Yanai Elazar and Valentin Hofmann and Ananya Harsh Jha and Sachin Kumar and Li Lucy and Xinxi Lyu and Nathan Lambert and Ian Magnusson and Jacob Morrison and Niklas Muennighoff and Aakanksha Naik and Crystal Nam and Matthew E. Peters and Abhilasha Ravichander and Kyle Richardson and Zejiang Shen and Emma Strubell and Nishant Subramani and Oyvind Tafjord and Pete Walsh and Luke Zettlemoyer and Noah A. Smith and Hannaneh Hajishirzi and Iz Beltagy and Dirk Groeneveld and Jesse Dodge and Kyle Lo},
      year={2024},
      eprint={2402.00159},
      archivePrefix={arXiv},
      primaryClass={cs.CL},
      url={https://arxiv.org/abs/2402.00159}, 
}

@misc{su2025nemotroncctransformingcommoncrawl,
      title={{Nemotron-CC}: Transforming {Common Crawl} into a Refined Long-Horizon Pretraining Dataset}, 
      author={Dan Su and Kezhi Kong and Ying Lin and Joseph Jennings and Brandon Norick and Markus Kliegl and Mostofa Patwary and Mohammad Shoeybi and Bryan Catanzaro},
      year={2025},
      eprint={2412.02595},
      archivePrefix={arXiv},
      primaryClass={cs.CL},
      url={https://arxiv.org/abs/2412.02595}, 
}

@article{elazar2023s,
  title={What's In My Big Data?},
  author={Elazar, Yanai and Bhagia, Akshita and Magnusson, Ian and Ravichander, Abhilasha and Schwenk, Dustin and Suhr, Alane and Walsh, Pete and Groeneveld, Dirk and Soldaini, Luca and Singh, Sameer and others},
  journal={arXiv preprint arXiv:2310.20707},
  year={2023}
}

@article{allen2023physics,
  title={Physics of language models: Part 3.1, knowledge storage and extraction},
  author={Allen-Zhu, Zeyuan and Li, Yuanzhi},
  journal={arXiv preprint arXiv:2309.14316},
  year={2023}
}

@article{allen2024physics,
  title={Physics of language models: Part 3.3, knowledge capacity scaling laws},
  author={Allen-Zhu, Zeyuan and Li, Yuanzhi},
  journal={arXiv preprint arXiv:2404.05405},
  year={2024}
}

@article{song2022learning,
  title     = {Learning from noisy labels with deep neural networks: A survey},
  author    = {Song, Hwanjun and Kim, Minseok and Park, Dongmin and Shin, Yooju and Lee, Jae-Gil},
  journal   = {IEEE transactions on neural networks and learning systems},
  volume    = {34},
  number    = {11},
  pages     = {8135--8153},
  year      = {2022},
  publisher = {IEEE}
}

@article{albalak2024survey,
  title   = {A survey on data selection for language models},
  author  = {Albalak, Alon and Elazar, Yanai and Xie, Sang Michael and Longpre, Shayne and Lambert, Nathan and Wang, Xinyi and Muennighoff, Niklas and Hou, Bairu and Pan, Liangming and Jeong, Haewon and others},
  journal = {arXiv preprint arXiv:2402.16827},
  year    = {2024}
}

@article{joulin2016fasttext,
  title   = {FastText.zip: Compressing text classification models},
  author  = {Joulin, Armand and Grave, Edouard and Bojanowski, Piotr and Douze, Matthijs and J{\'e}gou, H{\'e}rve and Mikolov, Tomas},
  journal = {arXiv preprint arXiv:1612.03651},
  year    = {2016}
}

@misc{hendrycks2019usingpretrainingimprovemodel,
  title         = {Using Pre-Training Can Improve Model Robustness and Uncertainty},
  author        = {Dan Hendrycks and Kimin Lee and Mantas Mazeika},
  year          = {2019},
  eprint        = {1901.09960},
  archiveprefix = {arXiv},
  primaryclass  = {cs.LG},
  url           = {https://arxiv.org/abs/1901.09960}
}

@misc{delétang2023neuralnetworkschomskyhierarchy,
  title         = {Neural Networks and the Chomsky Hierarchy},
  author        = {Grégoire Delétang and Anian Ruoss and Jordi Grau-Moya and Tim Genewein and Li Kevin Wenliang and Elliot Catt and Chris Cundy and Marcus Hutter and Shane Legg and Joel Veness and Pedro A. Ortega},
  year          = {2023},
  eprint        = {2207.02098},
  archiveprefix = {arXiv},
  primaryclass  = {cs.LG},
  url           = {https://arxiv.org/abs/2207.02098}
}

@inproceedings{zhou2024leveraging,
  title={Leveraging web-crawled data for high-quality fine-tuning},
  author={Zhou, Jing and Jiang, Chenglin and Shen, Wei and Zhou, Xiao and He, Xiaonan},
  booktitle={Findings of the Association for Computational Linguistics: EMNLP 2024},
  pages={11297--11312},
  year={2024}
}

@article{shinnick2025transformers,
  title={Transformers pretrained on procedural data contain modular structures for algorithmic reasoning},
  author={Shinnick, Zachary and Jiang, Liangze and Saratchandran, Hemanth and Hengel, Anton van den and Teney, Damien},
  journal={arXiv preprint arXiv:2505.22308},
  year={2025}
}
